\documentclass{article} 
\usepackage{iclr2027_conference,times}

\usepackage{amsmath,amsfonts,bm}

\def\eqref#1{equation~\ref{#1}}

\def\1{\bm{1}}

\DeclareMathAlphabet{\mathsfit}{\encodingdefault}{\sfdefault}{m}{sl}
\SetMathAlphabet{\mathsfit}{bold}{\encodingdefault}{\sfdefault}{bx}{n}

\usepackage{hyperref}
\usepackage{url}

\usepackage{booktabs}

\usepackage{amsmath}
\usepackage{amssymb}
\usepackage{amsfonts}
\usepackage{cleveref}
\usepackage{tabularx}
\usepackage{gensymb}

\usepackage{multirow}
\usepackage[table]{xcolor} 

\newcommand{\think}{\textsc{Think}}
\newcommand{\nothink}{\textsc{NoThink}}

\usepackage{tikz}
\usetikzlibrary{arrows.meta, positioning, calc}

\title{Thinking Leakage: A Causal Audit of NoThink Post-Training in Hybrid Reasoning Models}

\author{Zehao Liu, Vasant G. Honavar \\
  College of Information Sciences and Technology\\
  Pennsylvania State University \\
  \texttt{\{zml5418, vuh14\}@psu.edu}
}

\iclrfinalcopy 
\begin{document}
\addtocontents{toc}{\protect\setcounter{tocdepth}{-1}}

\maketitle

\begin{abstract}
Post-training hybrid reasoning models in \nothink{} mode has attracted growing interest as a way to improve performance while keeping inference fast. 
However, these gains may draw on thinking behavior already accessible through the base model's \think{} mode.
We formulate this \emph{thinking leakage} in a causal mediation framework and audit its contribution using bidirectional interventions along a simple base-derived activation direction.
Across three models and three post-training methods on competition math benchmarks, we find that leakage is real, causal, and
substantial: behavioral and representational analyses reveal shifts toward \think{}, steering the base model along this direction reproduces most of the post-training accuracy gain, and counter-steering a checkpoint removes a substantial share of what it gains.
Across nine aligned checkpoints with positive \nothink{} gains, the resulting leakage ratio ranges from $42\%$ to $79\%$.
These interventions support a substantial causal contribution of thinking leakage.
Our findings show that a post-training method's apparent advantage can therefore reflect greater drift toward \think{}, obscuring whether it improves capability within \nothink{} or more effectively
re-invokes existing \think{} behavior.
\end{abstract}

\section{Introduction}
Hybrid reasoning models use a single set of weights to support two inference modes (\think{} and \nothink{}) for problems of varying difficulty~\citep{yang2025qwen3, deepseekv32, gemma4, nvidia2026nemotron3nanoomni, kimiteam2026kimik25visualagentic}.
In \think{}, the model generates an explicit \textit{chain of thought} before responding to a prompt; in \nothink{}, it responds without the benefit of such a chain of thought.
Proprietary models from OpenAI and Anthropic also offer finer control through multiple levels of reasoning effort~\citep{openai_reasoning, anthropic_effort}.
Explicit reasoning can improve performance on challenging mathematical problems~\citep{snell2024scalingllmtesttimecompute}, whereas \nothink{} reduces inference cost~\citep{sui2025stop}.

Recent work has explored the post-training of hybrid models in the \nothink{} mode using reinforcement learning~\citep{zhu2025the, lin2026resrl, xu2026agpoasymmetricgrouppolicy, huang2026bootstrappingexplorationgrouplevelnatural}, self-distillation~\citep{hubotter2026reinforcement, li2026unifying, li2026onpolicyselfdistillationsupervision}, and on-policy distillation~\citep{yang2026learningteachergeneralizedonpolicy, ding2026doesonpolicydistillationreally}, reporting substantial gains on reasoning benchmarks.
However, this setting does not establish the source of the reported gains.

Post-training can elicit abilities that are latent in a base model~\citep{NEURIPS2025_537d5aa7, karan2026reasoning, NEURIPS2025_b8026fe0}. In hybrid reasoning models, however, strong reasoning abilities are already accessible through \think{} mode. Consistent with this distinction, checkpoints post-trained in \nothink{} mode still fall short of the base model’s \think{} performance when evaluated in \nothink{} (\cref{sec:correlational}). Thus, improved \nothink{} performance need not reflect reasoning capability newly elicited by post-training. We call this possibility \emph{thinking leakage}: a \nothink{} post-training gain that depends on increased engagement of reasoning behavior already accessible through the base model's \think{} mode.

Related behaviors have been observed in prior work~\citep{NEURIPS2025_b8026fe0, gan-etal-2026-thinking, zhang-etal-2025-adaptthink, NEURIPS2025_16371a9d}. Here, we ask whether \emph{thinking leakage} occurs during \nothink{} post-training and, if so, whether it merely correlates with performance gains or causally contributes to them—and by how much.

We formalize these questions with a causal model in which thinking leakage mediates the effect of post-training on \nothink{} accuracy, and operationalize leakage through a direction in the residual stream. Measurements of internal representations and generated behavior establish the presence of leakage, while interventions along this direction test its causal contribution. The resulting causal decomposition separates the observed gain into a component causally dependent on the audited leakage direction and a residual component, yielding the \emph{leakage ratio}. Across three models and three post-training methods, we find that this dependence is substantial: among nine aligned checkpoints with positive \nothink{} gains, the leakage ratio ranges from $42\%$ to $79\%$.

Our contributions are as follows.
\begin{itemize}
\item We formalize \emph{thinking leakage} as a causal mediator of \nothink{} post-training gains and operationalize it as a single direction in the residual stream along which post-training systematically displaces the \nothink{} state toward the \think{} state.

\item We establish the causal role of this mediator through bidirectional interventions. Moving base-model \nothink{} representations along the leakage direction reproduces much of the post-training gain, while reversing the displacement in post-trained models removes it, showing that a substantial share of the gain is mediated by reasoning behavior already accessible to the base model.
\item We derive a \emph{leakage ratio} that quantifies the fraction of each positive \nothink{} gain causally dependent on the audited leakage direction. Across the models, training methods, and checkpoints we audit, the leakage-dependent component scales approximately linearly with representational drift, with a single model-specific coefficient accounting for methods and training steps.

\end{itemize}

\section{A Causal Account of Thinking Leakage}
\label{sec:method}
\subsection{Problem Formulation}
\label{subsec:causal}
Hybrid models such as Qwen3~\citep{yang2025qwen3} and MiniCPM~\citep{minicpm4} expose \think{} and \nothink{} modes through the same parameters. We ask whether \nothink{} post-training gains arise partly from re-invoking computation already accessible in \think{}, and how much of the gain depends on that re-engagement.

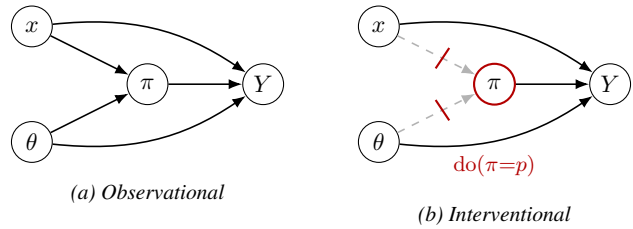
\begin{figure}[h!]
\centering
\begin{tikzpicture}[
  scale=0.9, transform shape,
  node/.style={circle, draw, minimum size=6mm, inner sep=1pt},
  edge/.style={-{Latex[length=1.8mm]}, semithick},
  cutedge/.style={-{Latex[length=1.8mm]}, semithick, gray!60, dashed},
  panel/.style={font=\small\itshape}
]
\begin{scope}[local bounding box=obs]
  \node[node] (x) at (0, 0.85) {$x$};
  \node[node] (theta) at (0, -0.85) {$\theta$};
  \node[node] (pi) at (1.7, 0) {$\pi$};
  \node[node] (y) at (3.4, 0) {$Y$};
  \draw[edge] (x) -- (pi);
  \draw[edge] (theta) -- (pi);
  \draw[edge] (pi) -- (y);
  \draw[edge] (theta) to[bend right=20] (y);
  \draw[edge] (x) to[bend left=20] (y);
\end{scope}
\node[panel, anchor=north] at ([yshift=-2mm]obs.south) {(a) Observational};

\begin{scope}[xshift=5.1cm, local bounding box=int]
  \node[node] (x2) at (0, 0.85) {$x$};
  \node[node] (theta2) at (0, -0.85) {$\theta$};
  \node[node, draw=red!70!black, thick] (pi2) at (1.7, 0) {$\pi$};
  \node[node] (y2) at (3.4, 0) {$Y$};
  \draw[cutedge] (x2) -- (pi2);
  \draw[cutedge] (theta2) -- (pi2);
  \draw[red!70!black, thick] ($(x2)!0.55!(pi2)+(-1mm,-1.5mm)$) -- ($(x2)!0.55!(pi2)+(1mm,1.5mm)$);
  \draw[red!70!black, thick] ($(theta2)!0.55!(pi2)+(-1mm,1.5mm)$) -- ($(theta2)!0.55!(pi2)+(1mm,-1.5mm)$);
  \draw[edge] (pi2) -- (y2);
  \draw[edge] (theta2) to[bend right=20] (y2);
  \draw[edge] (x2) to[bend left=20] (y2);
  \node[red!70!black, font=\footnotesize, anchor=north] at ([yshift=-6mm]pi2.south) {$\mathrm{do}(\pi{=}p)$};
\end{scope}
\node[panel, anchor=north] at ([yshift=-2mm]int.south) {(b) Interventional};
\end{tikzpicture}
\caption{Causal structure of a \nothink{} forward pass.
\textbf{(a)}~Observationally, thinking engagement $\pi$ is generated by the
input $x$ and weights $\theta$; accuracy $Y$ mixes the direct path
$\theta{\to}Y$ with the mediated path $\theta{\to}\pi{\to}Y$.
\textbf{(b)}~Steering and counter-steering intervene on $\pi$, changing its natural
value while leaving $\theta{\to}Y$ and $x{\to}Y$ intact: outcomes under
intervention still depend on the weights.}
\label{fig:causal-dag}
\end{figure}

Post-training can affect accuracy $Y$ directly and through engagement $\pi$ (\cref{fig:causal-dag}). Observation alone cannot separate these paths, so we intervene on $\pi$ while holding weights fixed. With $\theta_0,\theta_1$ denoting base and post-trained weights and $\pi_0,\pi_1$ their natural engagement levels, define
\begin{equation}
\begin{aligned}
Y_{0,0}&=Y(\theta_0,\pi_0),&Y_{0,1}&=Y(\theta_0,\pi_1),\\
Y_{1,0}&=Y(\theta_1,\pi_0),&Y_{1,1}&=Y(\theta_1,\pi_1).
\end{aligned}
\label{eq:four-outcomes}
\end{equation}
$Y_{0,0}$ and $Y_{1,1}$ are observed; steering estimates $Y_{0,1}$ and counter-steering estimates $Y_{1,0}$, enabling the decomposition in \cref{sec:gain_audit}.

\subsection{The Leakage Axis}
\label{subsec:directions}
Let $\bar h^{(\ell)}_{\mathcal M}(x;\theta)$ be the residual-stream activation at layer $\ell$, averaged over generated positions from mode $\mathcal M\in\{\mathcal T,\mathcal N\}$ and its rollouts for problem $x$, and $\bar h^{(\ell)}_{\mathcal M}(\theta)$ its dataset mean:
\begin{equation}
\bar h^{(\ell)}_{\mathcal M}(x;\theta)=\frac{1}{|\mathcal M(x)|}\sum_{t\in\mathcal M(x)}h_t^{(\ell)}(x;\theta),\quad
\bar h^{(\ell)}_{\mathcal M}(\theta)=\mathbb E_x[\bar h^{(\ell)}_{\mathcal M}(x;\theta)].
\label{eq:hbar}
\end{equation}
Following difference-of-means approaches~\citep{marks2023geometry, arditi2024refusal}, we define the \emph{leakage vector} and \emph{leakage axis} as
\begin{equation}
v^{(\ell)}_{\rm leak}=\bar h^{(\ell)}_{\mathcal T}(\theta_0)-\bar h^{(\ell)}_{\mathcal N}(\theta_0),\qquad
\hat v^{(\ell)}_{\rm leak}=v^{(\ell)}_{\rm leak}/\|v^{(\ell)}_{\rm leak}\|.
\label{eq:vleak}
\end{equation}
We operationalize engagement as the \nothink{} projection
\begin{equation}
\pi^{(\ell)}(x;\theta)=\langle\bar h^{(\ell)}_{\mathcal N}(x;\theta),\hat v^{(\ell)}_{\rm leak}\rangle,\qquad
\pi^{(\ell)}_\theta=\mathbb E_x[\pi^{(\ell)}(x;\theta)],
\label{eq:pi}
\end{equation}
and normalize checkpoint drift by the base \think{}--\nothink{} gap:
\begin{equation}
\Delta\pi=(\pi^{(\ell)}_1-\pi^{(\ell)}_0)/\|v^{(\ell)}_{\rm leak}\|.
\label{eq:delta_pi}
\end{equation}
Thus base \nothink{} is $0$ and base \think{} lies at $1$ on the normalized displacement scale. Unless noted otherwise, we audit layer $\ell=20$.

\subsection{Experimental Setup}
\label{sec:exp_setup}
We audit Qwen3-8B, Qwen3-4B~\citep{yang2025qwen3}, and MiniCPM4.1-8B~\citep{minicpm4}, post-trained only in \nothink{} using GRPO~\citep{guo2025deepseek, shao2024deepseekmath}, SFT, or OPSD with or without LoRA~\citep{zhao2026self}. Evaluation uses $120$ competition-math problems, $30$ each from AIME24~\citep{aime24}, AIME25~\citep{aime25}, HMMT-Feb-2025, and HMMT-Nov-2025~\citep{dekoninck2026matharena}. Full training, sampling, intervention, and evaluation details are in \cref{app:setup}.

\section{Does Leakage Occur? Detecting Drift in \nothink{}}
\label{sec:correlational}
We first test whether post-training shifts \nothink{} toward \think{}. At layer $20$ we measure: (i) alignment $c_s=\cos(\delta_s,\hat v_{\rm leak})$ of checkpoint drift $\delta_s=\bar h_{\mathcal N}(\theta_s)-\bar h_{\mathcal N}(\theta_0)$; (ii) problem-level paired projection drift summarized by repeated-measures Cohen's $d$; (iii) density of \think{}-characteristic reasoning markers; and (iv) the checkpoint-specific \think{}--\nothink{} mode gap. Formally,
\begin{equation}
c_s=\frac{\pi_s-\pi_0}{\|\delta_s\|},\qquad
d_s=\frac{\overline{\Delta}}{s_\Delta},\quad
\Delta_q=\pi(x_q;\theta_s)-\pi(x_q;\theta_0),
\label{eq:align}
\end{equation}
and marker density is
\begin{equation}
b(x)=\frac{\#\{\text{reasoning markers in }x\}}{\#\{\text{generated tokens in }x\}}.
\label{eq:marker-density}
\end{equation}
For geometry we recompute the two mode centroids within each checkpoint:
\begin{equation}
v_{\rm leak}(\theta)
=\bar h_{\mathcal T}(\theta)-\bar h_{\mathcal N}(\theta),
\qquad
\phi(\theta)
=\angle\!\left(\bar h_{\mathcal T}(\theta),\bar h_{\mathcal N}(\theta)\right).
\label{eq:modegap}
\end{equation}
Layerwise analyses are in \cref{app:vref-depth,app:geometry}.

\begin{figure*}[h]
    \centering
    \includegraphics[width=\textwidth]{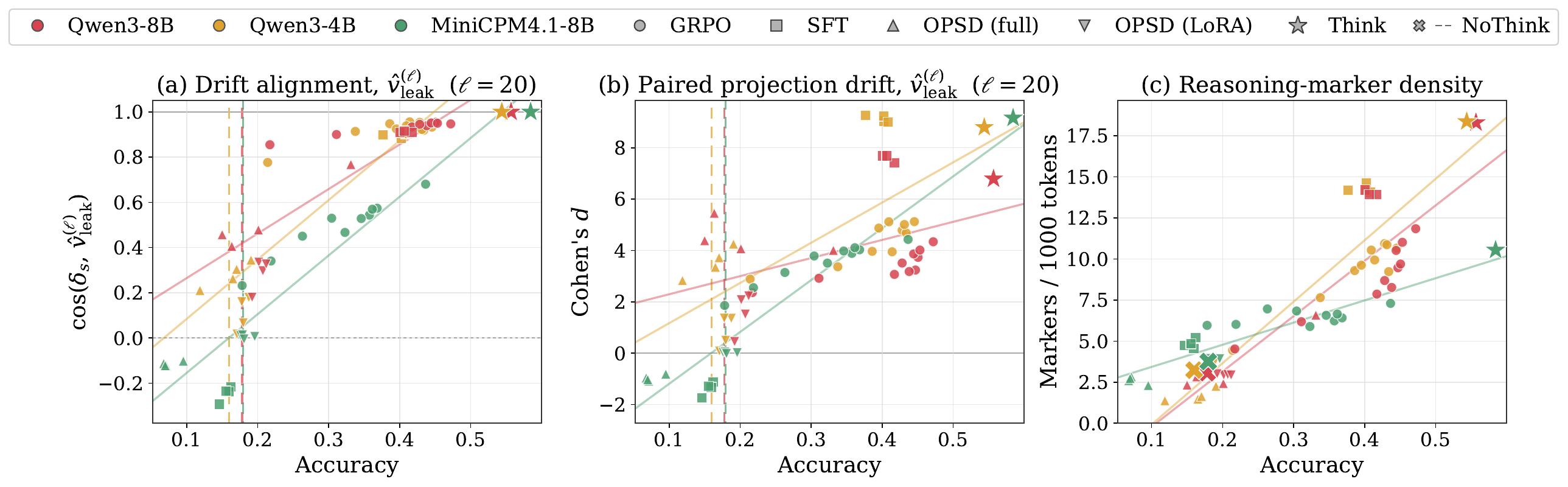}
    \caption{%
    \textbf{Thinking leakage tracks \nothink{} accuracy across checkpoints of different training methods and model scales.} Each point is one checkpoint; marker shape denotes the training method and color the base model. $\times$ and $\star$ are the base model in \nothink{} and \think{}.
    \textbf{(a)} Drift alignment to the leakage axis $\hat{v}_{\mathrm{leak}}$.
    \textbf{(b)} Paired projection drift (Cohen's $d$) along $\hat{v}_{\mathrm{leak}}$.
    \textbf{(c)} Reasoning-marker density in the \nothink{} rollouts.
    More experiments and analysis are given in \cref{app:leakage}
}
    \label{fig:leakage-main}
\end{figure*}

\paragraph{Leakage occurs, and it scales with the reported gain.}
\Cref{fig:leakage-main} plots the first three measures against \nothink{}
accuracy across all checkpoints. All three views---population-level,
problem-level, and behavioral---agree: as \nothink{} accuracy improves,
representations drift toward the \think{} state and generated responses
become more \think{}-like ($\rho=+0.92$, $+0.68$, and $+0.85$,
respectively). The highest-accuracy checkpoints exhibit the strongest
\think{}-ward drift, while the few checkpoints that lose accuracy
relative to base move in the opposite direction along the leakage axis.
Thus, across checkpoints, improvements in \nothink{} accuracy are tightly
coupled to increased engagement of the \think{}-associated direction.

The coupling is not, however, fixed across training methods. At matched
\nothink{} accuracy on the Qwen models, SFT moves more than twice as far
along $\hat v_{\mathrm{leak}}$ as GRPO (\cref{app:methods}). Accuracy
alone therefore does not determine the amount of leakage, motivating the
causal interventions that follow.

\paragraph{The drift is specific to the reasoning-associated direction.}
A generic consequence of post-training---for example, learning to produce
better direct answers---could also shift internal representations. We
therefore repeat the representational analyses using a reference axis
defined by the clean solution that a \think{} rollout produces
\emph{after} deliberation, rather than by the deliberative portion itself
(\cref{app:vref}). The contrast is sharp: the effect size along the
leakage axis is more than an order of magnitude larger ($d=3.15$ versus
$0.02$), and only drift along the leakage axis strongly tracks accuracy
($\rho=+0.68$ versus $+0.23$). This separation persists at every probed
layer (\cref{app:vref-depth}). The observed drift therefore aligns
specifically with the representation associated with deliberation, rather
than with a generic shift toward better final-answer representations.

\begin{figure}[h]
    \centering
    \includegraphics[width=0.65\linewidth]{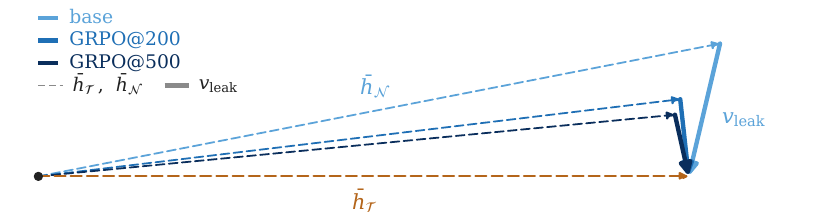}
    \caption{\textbf{GRPO collapses the mode gap.}
Layer-20 geometry of Qwen3-8B at GRPO steps $0$, $200$, $500$.
$\bar h_{\mathcal T}$, $\bar h_{\mathcal N}$ (dashed), and $v_{\mathrm{leak}}=\bar h_{\mathcal T}-\bar h_{\mathcal N}$ (solid) are recomputed within each checkpoint.
Both $\bar h_{\mathcal T}$, $\bar h_{\mathcal N}$ keep their norms to within a few percent, yet the angle between them closes from $11.31^\circ$ to $5.48^\circ$ and $\|v_{\mathrm{leak}}\|$ falls $15.43 \to 8.81 \to 7.18$.
The panel is an approximate embedding rather than an orthogonal projection of $4096$-d quantities; embedding errors are reported in \cref{app:geometry}.}
    \label{fig:vleak-geometry-a}
\end{figure}

\paragraph{\nothink{} converges toward \think{}.}
The geometry of the two modes provides a complementary view. Recomputing
their centroids within each checkpoint shows that GRPO reduces the
\think{}--\nothink{} gap by more than half on Qwen3-8B
(\cref{fig:vleak-geometry-a}). The norms of both
$\bar h_{\mathcal T}$ and $\bar h_{\mathcal N}$ remain within a few
percent of their base values, while $84\%$ of the reduction in the gap
is attributable to the narrowing angle between them.

This convergence is strongly asymmetric. Measured relative to the base
mode gap, $\bar h_{\mathcal T}$ moves only $5.8\%$ and $10.2\%$ at the
two checkpoints, whereas $\bar h_{\mathcal N}$ moves $53.2\%$ and
$66.2\%$. Thus, the shrinking separation is driven primarily by
\nothink{} moving toward a comparatively stable \think{} representation,
rather than by both modes moving toward a new common state. The same
ordering holds at every probed layer (\cref{app:geometry}).

These results establish that leakage accompanies successful post-training,
but not that the displacement causes the gain. We test that claim next by
intervening directly on the leakage direction.

\section{Is the Direction Sufficient? Inducing Leakage by Steering}
\label{sec:sufficiency}

\Cref{sec:correlational} shows that \nothink{} post-training moves
representations along $\hat v_{\mathrm{leak}}$, but correlation alone
cannot establish whether this displacement contributes to the accompanying
behavior. We therefore test its causal \emph{sufficiency}: if we move the
untouched base model's \nothink{} state toward \think{} along
$\hat v_{\mathrm{leak}}$, without changing its weights, can we reproduce
the behavioral effects associated with post-training? In the causal model,
this amounts to estimating
\[
Y\big(\theta_0,\mathrm{do}(\pi{=}p)\big)
\]
over a range of engagement levels $p$ and comparing it with the untreated
baseline $Y(\theta_0,\pi{=}\pi_0)$.

We implement this intervention using additive activation
steering~\citep{rimsky2024steering}. During \nothink{} generation, we add
a scaled copy of the leakage vector at every generated token position at
layer $20$---the layer at which $\pi$ is measured and a depth at which
representations have been found to be causally manipulable by activation
interventions~\citep{zou2025representationengineeringtopdownapproach,
rimsky2024steering, NEURIPS2024_58cbe393}:
\begin{equation}
h_t
\;\leftarrow\;
h_t+\alpha v_{\mathrm{leak}}
\qquad\Longrightarrow\qquad
\pi
\;\mapsto\;
\pi+\alpha\lVert v_{\mathrm{leak}}\rVert
\;=:\;p_\alpha,
\qquad \alpha>0.
\label{eq:steer}
\end{equation}
Because $v_{\mathrm{leak}}$ spans the base model's mean
\nothink{}--\think{} separation, $\alpha$ is measured in units of that
gap: $\alpha=1$ adds one full base-mode displacement along the leakage
axis.

We sweep $\alpha\in\{0.5,1,1.25,1.5,2\}$ and measure how accuracy and
reasoning-marker density change as the base \nothink{} state is displaced
toward \think{}. If steering alone moves both outcomes toward those of the
post-trained checkpoints, then displacement along the leakage direction is
causally sufficient to induce the corresponding behavioral changes. This
forward intervention is paired with the counter-steering experiment in
\cref{sec:necessity}, which reverses the same displacement to test whether
it is also necessary for the observed post-training gain.

\begin{figure}[t]
    \centering
    \includegraphics[width=0.5\linewidth]{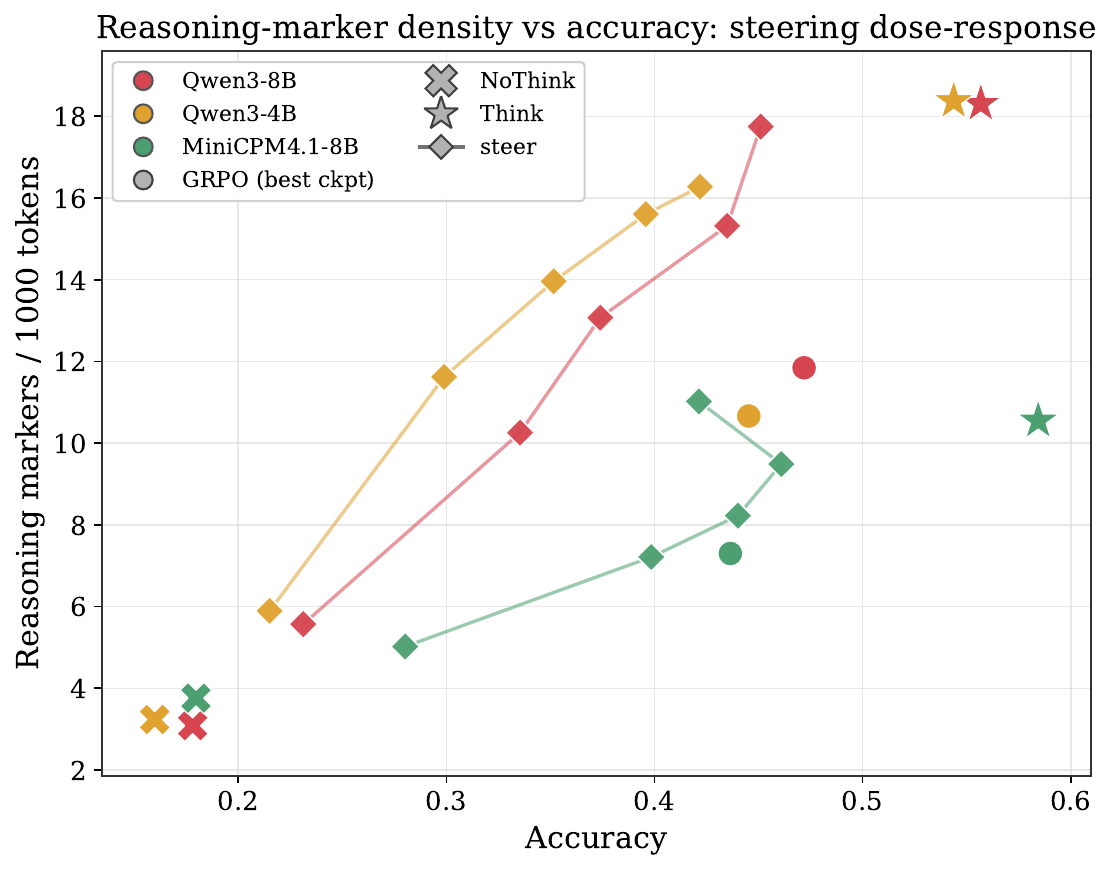}
    \caption{\textbf{Steering along $\hat{v}_{\mathrm{leak}}$ recovers the \think{} performance without training.} Reasoning-marker density versus \nothink{} accuracy, coloured by base model. Diamonds trace the layer-$20$ induction sweep over five doses; $\times$ and $\star$ are the unsteered base model in \nothink{} and \think{}, and the circle is each model's best post-training checkpoint.
}
    \label{fig:steer}
\end{figure}

\paragraph{One direction reproduces most of the post-training gain.}
\Cref{fig:steer} plots the five-dose sweep in the
accuracy--marker-density plane. As $\alpha$ increases, the base model
moves smoothly up and to the right from its \nothink{} baseline toward
its own \think{} point. Without any weight update, the best steered point
recovers $92$--$110\%$ of the best checkpoint's accuracy gain over base
\nothink{}, and roughly $70\%$ of the full \nothink{}$\to$\think{}
accuracy gap on all three models. On MiniCPM4.1-8B, steering surpasses
the best checkpoint; on both Qwen models, accuracy is still increasing at
the largest dose tested.

This effect is specific to the leakage direction: applying the same
intervention along a reference axis yields no comparable gain
(\cref{app:vref-steer}), while steering along
$\hat v_{\mathrm{leak}}$ at other depths produces the same qualitative
trend (\cref{app:layer-sweep}). Thus, displacement along a single
base-model direction is causally sufficient to reproduce most of the
observed post-training accuracy gain.

The axis is not exhausted by post-training. Applying the same intervention
to an already post-trained checkpoint yields further improvement:
steering Qwen3-8B's best GRPO checkpoint with $\alpha=1$ raises
accuracy from $0.472$ to $0.553$, within $0.4$ percentage points of base
\think{} performance (\cref{app:steer-trained}).

\paragraph{Steering induces more reasoning markers than training.}
Steering and post-training do not, however, produce identical behavior.
At its best dose, the steered model reaches $89$--$97\%$ of the base
\think{} marker density, whereas the best post-trained checkpoint reaches
only $58$--$69\%$. Per point of accuracy gained, steering produces
$1.5$--$1.9\times$ as many reasoning markers as post-training. Thus,
direct activation injection drives the surface signature associated with
\think{} more strongly than training does, while converting that shift
into accuracy less efficiently. The intervention therefore reproduces
most of the accuracy gain without simply reproducing the checkpoint's
entire behavioral profile.

\paragraph{At the largest dose, token budget becomes limiting.}
MiniCPM4.1-8B is the only model whose accuracy declines at $\alpha=2$,
falling $4.0$ percentage points from its peak at $\alpha=1.5$ even as
reasoning-marker density continues to rise. At this dose, steering also
lengthens responses sufficiently that they more often reach the token
limit (\cref{app:degrade}). The decline is therefore consistent with a
generation-budget constraint rather than a reversal of the steering
effect, and does not affect the sufficiency result at lower doses.

Sufficiency alone, however, does not show that post-trained checkpoints
actually depend on the displacement they acquire. Counter-steering tests
that complementary claim.

\section{Does the Gain Depend on Leakage? Removing It by Counter-Steering}
\label{sec:necessity}

\Cref{sec:sufficiency} shows that displacement along the leakage direction
is sufficient to reproduce most of the post-training accuracy gain. We now
ask the complementary question: how much of the observed gain depends on
the displacement that post-training actually induces? We
\emph{counter-steer} a post-trained checkpoint $\theta_1$, moving its
\nothink{} state back along $v_{\mathrm{leak}}$ toward base engagement
and measuring how much of the gain remains. In the causal model, this
estimates
\[
Y\big(\theta_1,\mathrm{do}(\pi{=}p)\big)
\]
at engagement levels $p<\pi_1$, relative to the untreated outcome
$Y(\theta_1,\pi{=}\pi_1)$.

By \cref{eq:delta_pi}, $\Delta\pi$ is the checkpoint's normalized
engagement drift from base \nothink{}. We reverse the steering
intervention of \cref{eq:steer}, scaling it by the drift measured for each
checkpoint. During \nothink{} generation, at layer $20$ we apply
\begin{equation}
h_t
\;\leftarrow\;
h_t-\gamma\,\Delta\pi\,v_{\mathrm{leak}}
\qquad\Longrightarrow\qquad
\pi
\;\mapsto\;
\pi-\gamma\,\Delta\pi\,\lVert v_{\mathrm{leak}}\rVert
\;=:\;p_\gamma,
\qquad \gamma>0.
\label{eq:counter-steer}
\end{equation}
The dose is therefore expressed in units of the checkpoint's own drift.
Our primary intervention uses $\gamma=1$, which exactly removes the
checkpoint's acquired displacement along the leakage axis and, by
construction, returns its mean projected engagement $\pi$ to the base
\nothink{} level. Any resulting loss in accuracy measures the extent to
which the checkpoint's gain causally depends on that displacement; the
corresponding change in reasoning-marker density provides a behavioral
check. This reversal forms the basis of the gain decomposition in
\cref{sec:gain_audit}.

\paragraph{Checkpoint selection.}
For each model, we counter-steer the best checkpoint from each
post-training method (\textit{GRPO/SFT/OPSD best}), an intermediate
\textit{GRPO@200} checkpoint, and the most severely collapsed OPSD
checkpoint, included to test the intervention under reverse drift as well
as successful training. Based on performance and drift alignment
(\cref{sec:correlational}), the fifteen checkpoints fall into three groups
(\cref{app:checkpoint-groups}). Here we focus on the nine
\emph{aligned} checkpoints, which improve over base while drifting toward
\think{}, and the two \emph{reverse-drift} checkpoints, which lose
accuracy while moving in the opposite direction. The remaining
checkpoints and complete grouping criteria are reported in
\cref{app:checkpoint-groups}.
\begin{figure}[t]
\centering
\includegraphics[width=0.5\linewidth]{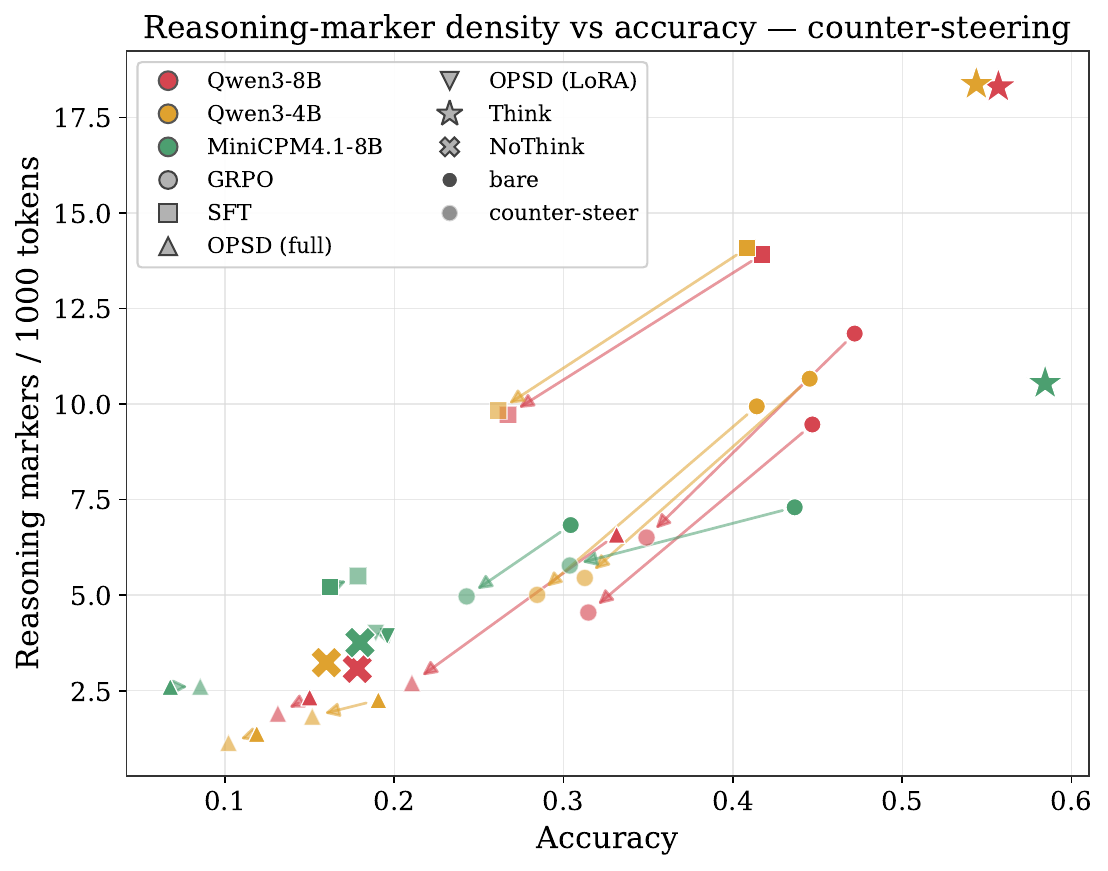}
\caption{\textbf{Counter-steering along $\hat{v}_{\mathrm{leak}}$ removes the post-training gains.}
Reasoning-marker density versus \nothink{} accuracy, coloured by base model. Each checkpoint is a before $\to$ after pair joined by an arrow; $\times$ and $\star$ are the unsteered base model in \nothink{} and \think{}.
}
\label{fig:marker_vs_acc_counter-steer}
\end{figure}

\paragraph{Counter-steering removes much of the post-training gain.}
Across the nine aligned checkpoints, counter-steering reverses a substantial
fraction of both the accuracy and reasoning-marker gains acquired during
post-training (\cref{fig:marker_vs_acc_counter-steer}). Relative to each
checkpoint's improvement over base \nothink{}, counter-steering removes,
on average, $54.5\%$ of the accuracy gain and $63.8\%$ of the
marker-density gain. The largest reversal occurs for Qwen3-8B OPSD@25,
where $78.9\%$ of the accuracy gain is removed. Per-checkpoint results
are reported in \cref{tab:counter-steer-full}.

\paragraph{The effect follows the sign of the learned drift.}
The two reverse-drift checkpoints provide a signed control. Because their
post-training displacement along $v_{\mathrm{leak}}$ is negative relative
to base \nothink{}, the same $\gamma=1$ intervention adds
$v_{\mathrm{leak}}$ rather than subtracting it. Their accuracy
\emph{increases}. Thus, across both aligned and reverse-drift checkpoints,
counter-steering consistently reverses the displacement induced by
post-training: it removes accuracy when training moves the model toward
\think{}, and restores accuracy when training moves it away. This
sign-sensitive reversal is difficult to explain as generic intervention
damage, which would instead predict degradation in both cases.
Consistent with this interpretation, matched interventions along a random
direction or the reference axis $v_{\mathrm{ref}}$ leave accuracy and
reasoning-marker density essentially unchanged
(\cref{app:counter-steer-controls}).

\paragraph{Counter-steering moves behavior back toward base.}
Counter-steering also increases degenerate repetition, raising an
alternative explanation: perhaps the intervention lowers accuracy simply
by damaging generation rather than by removing leakage. Two checks argue
against this account. First, at the problem level, the induced repetition
tracks the base model's own repetition pattern, while the checkpoint's
accuracy pattern shifts toward that of the base model as the intervention
dose increases. Second, restricting the analysis to rollouts that
terminate normally preserves $91\%$ of the observed accuracy loss.
Thus, only a small fraction of the reversal can be attributed to
degenerate generation; most persists among normally terminating responses
(\cref{app:counter-steer-reversion}).

\section{How Much of the Gain Is Thinking Leakage?}
\label{sec:gain_audit}
The preceding experiments establish both directions of the causal test:
moving the base model forward along the leakage axis improves performance,
while removing acquired displacement from post-trained checkpoints reduces
it. We now combine these interventions to quantify how much of each
observed gain depends on that displacement.

Of the four potential outcomes in \cref{eq:four-outcomes}, $Y_{0,0}$ and
$Y_{1,1}$ are observed directly. Counter-steering yields $Y_{1,0}$;
steering the base by the same normalized displacement yields $Y_{0,1}$,
interpolated at $\alpha^\star=\Delta\pi$. The observed gain has the exact decomposition
\begin{equation}
\Delta=Y_{1,1}-Y_{0,0}
=\underbrace{(Y_{1,1}-Y_{1,0})}_{L_{\rm trained}}
+\underbrace{(Y_{1,0}-Y_{0,0})}_{R_0}.
\label{eq:trained-decomp}
\end{equation}
$L_{\rm trained}$ is the accuracy lost when acquired leakage displacement is removed; $R_0$ is what remains after projected engagement is restored to base.

\begin{figure}[t]
\centering
\begin{minipage}[c]{0.46\linewidth}
\centering
\begin{equation}
\begin{aligned}
\Delta
&=
\underbrace{Y_{1,1}-Y_{1,0}}_{L_{\mathrm{trained}}}
+
\underbrace{Y_{1,0}-Y_{0,0}}_{R_0}\\
&=
\underbrace{Y_{0,1}-Y_{0,0}}_{L_{\mathrm{base}}}
+
\underbrace{Y_{1,1}-Y_{0,1}}_{R_1}.
\end{aligned}
\label{eq:bidirectional_gain_decomposition}
\end{equation}
\end{minipage}
\hfill
\begin{minipage}[c]{0.5\linewidth}
\centering
\includegraphics[width=0.75\linewidth]{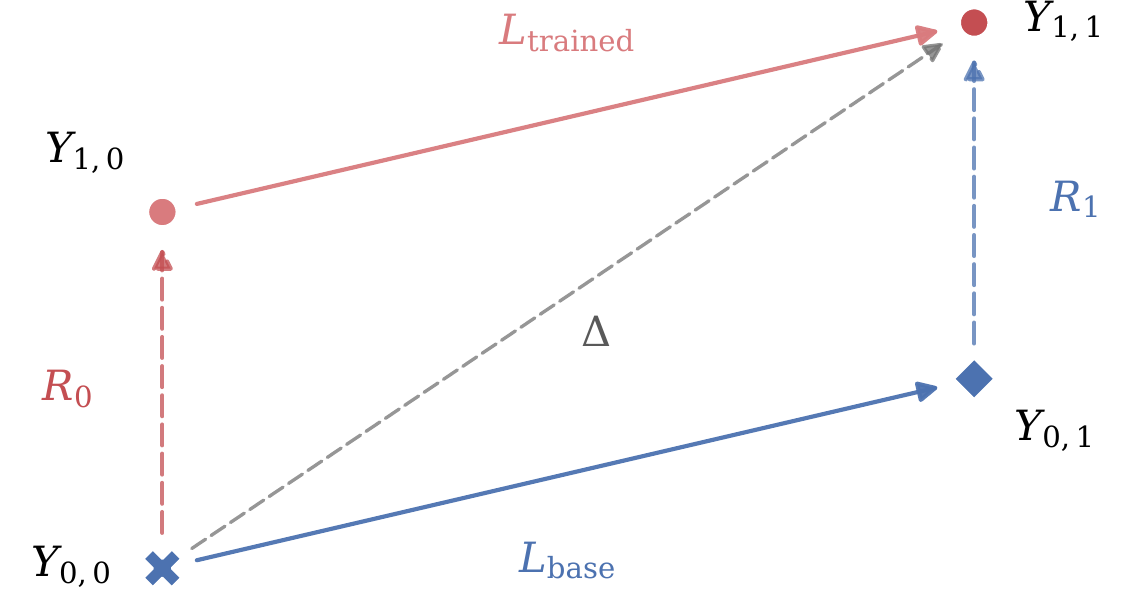}
\end{minipage}
\caption{\textbf{The two decompositions of the total gain.}
Left: the algebraic decomposition of $\Delta$ into two paths, $(L_{\mathrm{trained}}, R_0)$ and $(L_{\mathrm{base}}, R_1)$. 
Right: a schematic illustration of the same decomposition.
}
\label{fig:decomp-schematic}
\end{figure}
\begin{figure}[t]
\centering
\includegraphics[width=1\linewidth]{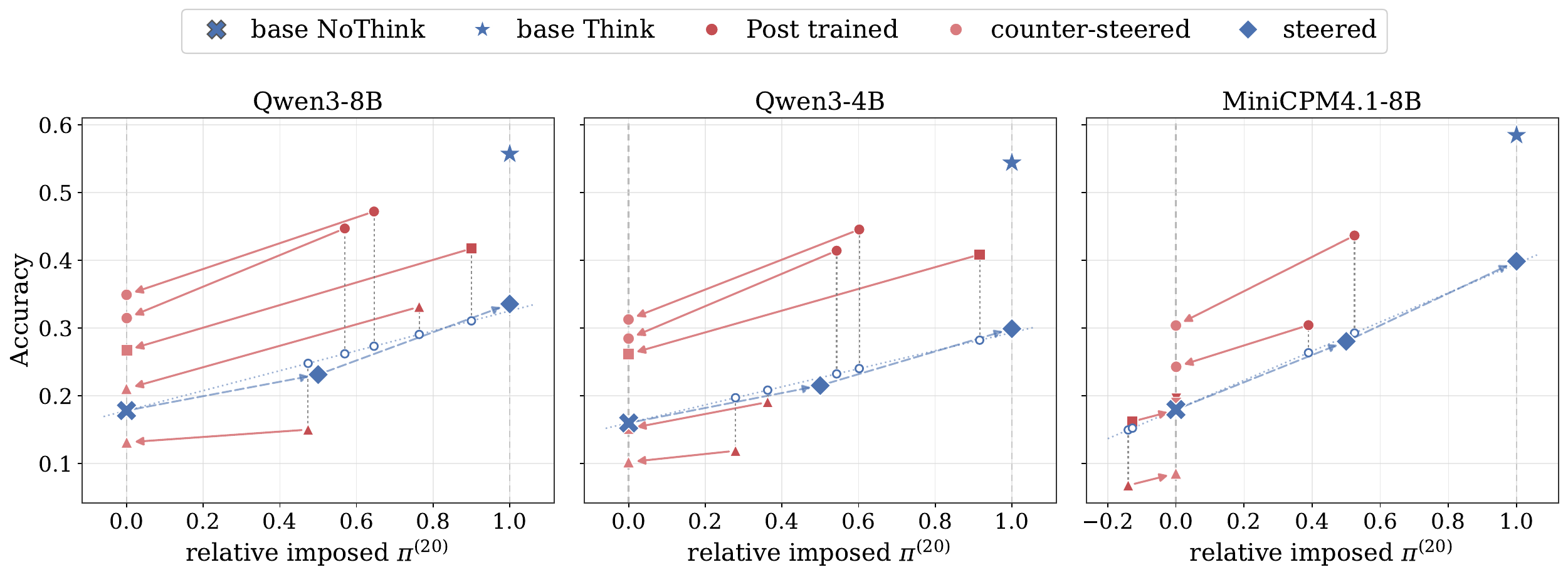}
\caption{\textbf{\nothink{} accuracy against the engagement}, at layer 20, one panel per model. Blue: the base model under steering, tracing $Y_{0,0}\!\to\!Y_{0,1}$. Red: each post-trained checkpoint and its counter-steered counterpart, tracing $Y_{1,1}\!\to\!Y_{1,0}$. Arrows show the direction of the intervention, not of the decomposition.
$L_{\mathrm{trained}} = Y_{1,1} - Y_{1,0}$ is the height the checkpoint loses along the red arrow, and $R_0 = Y_{1,0} - Y_{0,0}$ compares the counter-steered checkpoint with the untreated base.
}
\label{fig:profiles}
\end{figure}

The complementary path through $Y_{0,1}$ (the second path in \cref{fig:decomp-schematic}) asks the same question from the
untouched base model: how much accuracy is induced when the base is moved
forward by exactly the displacement acquired during training. Agreement
between these two paths therefore provides a check that the estimated
leakage contribution is not peculiar to either steering the base or
counter-steering the trained checkpoint.

\paragraph{Leakage scales linearly with drift.}
Across all 15 checkpoints, $L_{\rm trained}$ is approximately linear in $\Delta\pi$ within each model ($R^2=0.92$--$0.95$, \cref{fig:profiles}):
\begin{equation}
L_{\rm trained}\approx\beta_{\rm trained}\Delta\pi,\qquad
\Delta\approx\beta_{\rm trained}\Delta\pi+R_0,
\label{eq:linear-decomp}
\end{equation}
with $\beta_{\rm trained}=0.166,0.178,0.212$ for Qwen3-8B, Qwen3-4B, and MiniCPM4.1-8B. Thus one model-specific coefficient predicts the leakage-dependent component across methods and training steps. In contrast, $R_0$'s association with drift falls from $\rho=+0.69$ to $-0.20$ after controlling for total gain, while its association with total gain remains $\rho=+0.97$ after controlling for drift. A method-level comparison using this decomposition is deferred to \cref{app:methods-decomposition}.

\paragraph{The opposite intervention agrees.}
The complementary path gives
\begin{equation}
\Delta=L_{\rm base}+R_1,
\qquad
L_{\rm base}\approx\beta_{\rm base}\Delta\pi,
\label{eq:base-decomp}
\end{equation}
with $\beta_{\rm base}=0.147,0.133,0.215$. The ratios $\beta_{\rm trained}/\beta_{\rm base}$ are $1.13,1.34,0.99$, and the checkpoint-level asymmetry
\begin{equation}
\mathcal A=L_{\rm trained}-L_{\rm base}=R_1-R_0
\label{eq:bidirectional_asymmetry}
\end{equation}
averages only $2.6$ accuracy points in absolute value. Base steering and checkpoint counter-steering therefore assign similar accuracy changes to the same displacement; problem-level agreement is in \cref{app:question-level}.

\paragraph{Leakage ratio.}
For positive-gain checkpoints, define
\begin{equation}
\lambda=\frac{L_{\rm trained}}{\Delta}
=\frac{Y_{1,1}-Y_{1,0}}{Y_{1,1}-Y_{0,0}}.
\label{eq:leakage-ratio}
\end{equation}
Using $L_{\rm trained}\approx\beta_{\rm trained}\Delta\pi$,
this is equivalently
$\lambda\approx\beta_{\rm trained}\Delta\pi/\Delta$.
Across the nine aligned checkpoints, $\lambda=0.42$--$0.79$: roughly $42\%$--$79\%$ of observed improvement depends on the acquired displacement. GRPO lies near $0.5$, SFT near $0.6$, and Qwen3-8B OPSD@25 reaches $0.79$. The additive decomposition also applies to reverse-drift and degrading checkpoints, but for those we retain signed components rather than interpret $\lambda$ as a share of positive gain (\cref{app:detail_decomposition}).

\section{Related Work}
\label{sec:related}

\paragraph{Steering model behavior.}
Activation directions have been used to control traits including
sycophancy, honesty, refusal, and hallucination~\citep{rimsky2024steering,
zou2025representationengineeringtopdownapproach,NEURIPS2024_58cbe393,
chen2025personavectorsmonitoringcontrolling,pmlr-v267-wang25i}, and
reasoning behaviors such as thinking speed and backtracking
~\citep{NEURIPS2025_70d46c12,venhoff2025understandingreasoningthinkinglanguage,
sinii-etal-2025-steering}. Unlike work requiring targeted extraction, our
native \think{}--\nothink{} contrast yields a simple difference-of-means
direction used to audit post-training rather than merely control behavior.

\paragraph{Causal interventions in model interpretability.}
Internal interventions have been used to study mediation of gender
bias~\citep{vig2020investigating}, factual recall~\citep{meng2022locating},
and task-specific computations~\citep{wang2023interpretability}; their
interpretation depends on intervention design and evaluation
~\citep{zhang2024towards}. We use interventions in both directions to
quantify how much \nothink{} gain depends on a targeted
reasoning-associated direction. Extended related work is in
\cref{app:related}.

\section{Conclusion}
\label{sec:conclusion}
We investigate whether reported \nothink{} post-training gains depend on \emph{thinking leakage}: increased engagement of reasoning behavior already accessible through the base model’s \think{} mode. 
Across three hybrid reasoning models and three post-training methods, \nothink{} improvements are accompanied by drift toward \think{} in representation and behavior. A base-derived direction both reproduces most of the gain when injected and removes a substantial fraction when reversed. Among nine positive-gain aligned checkpoints, $42\%$--$79\%$ of improvement disappears when projected engagement is restored to base \nothink{}, and the amount removed scales approximately linearly with drift across methods and training steps.

Thus disabling explicit thinking at the interface does not ensure that post-training preserves the intended mode separation: apparent \nothink{} gains can partly recruit reasoning behavior already accessible through \think{}. Auditing such methods therefore requires asking not only how much accuracy improves, but what the improvement depends on (\cref{app:extended-discussion1}).

\section{Limitations and Future Work}
\label{sec:limitations}
Our experiments cover three models, three post-training methods, and competition mathematics; leakage may differ across architectures, scales, training procedures, and tasks. Broader coverage would clarify the generality of our findings.
Moreover, a single base-derived direction need not capture every form of leakage or remain complete after large model changes. Accordingly, $\lambda$ measures dependence on the audited component, not all possible leakage, and the residual $R_0$ cannot be identified as wholly leakage-independent or newly acquired \nothink{} capability. Multidimensional or nonlinear interventions could improve coverage and specificity (\cref{app:extended-discussion2}).

\subsection*{AI Use Statement}
The authors conceptualized and executed the study. Generative AI was not
used to generate synthetic datasets, perform the experimental evaluations
or statistical analyses, formulate mathematical claims, or assist with
proofs. It was used to critique aspects of the methodology, experimental
setup, and interpretation; assist with coding and figures; summarize,
analyze, brainstorm, and retrieve relevant literature; and edit or
rephrase text for clarity. All AI-assisted work and code were reviewed by
the authors, and all citations were manually verified and added by the
authors. The authors accept full responsibility for the manuscript and
artifacts.

\subsection*{Reproducibility Statement}
All models and benchmarks are public. Post-training and interventions use public frameworks; full details are in \cref{app:setup}. Code will be released upon publication.

\bibliography{iclr2027_conference}

@article{yang2025qwen3,
  title={Qwen3 technical report},
  author={An Yang and Anfeng Li and Baosong Yang and Beichen Zhang and Binyuan Hui and Bo Zheng and Bowen Yu and Chang Gao and Chengen Huang and Chenxu Lv and Chujie Zheng and Dayiheng Liu and Fan Zhou and Fei Huang and Feng Hu and Hao Ge and Haoran Wei and Huan Lin and Jialong Tang and Jian Yang and Jianhong Tu and Jianwei Zhang and Jianxin Yang and Jiaxi Yang and Jing Zhou and Jingren Zhou and Junyang Lin and Kai Dang and Keqin Bao and Kexin Yang and Le Yu and Lianghao Deng and Mei Li and Mingfeng Xue and Mingze Li and Pei Zhang and Peng Wang and Qin Zhu and Rui Men and Ruize Gao and Shixuan Liu and Shuang Luo and Tianhao Li and Tianyi Tang and Wenbiao Yin and Xingzhang Ren and Xinyu Wang and Xinyu Zhang and Xuancheng Ren and Yang Fan and Yang Su and Yichang Zhang and Yinger Zhang and Yu Wan and Yuqiong Liu and Zekun Wang and Zeyu Cui and Zhenru Zhang and Zhipeng Zhou and Zihan Qiu},
  journal={arXiv preprint arXiv:2505.09388},
  year={2025}
}

@article{minicpm4,
  title={Minicpm4: Ultra-efficient llms on end devices},
  author={MiniCPMTeam and Chaojun Xiao and Yuxuan Li and Xu Han and Yuzhuo Bai and Jie Cai and Haotian Chen and Wentong Chen and Xin Cong and Ganqu Cui and Ning Ding and Shengda Fan and Yewei Fang and Zixuan Fu and Wenyu Guan and Yitong Guan and Junshao Guo and Yufeng Han and Bingxiang He and Yuxiang Huang and Baoxi Ji and Cunliang Kong and Qiuzuo Li and Siyuan Li and Wenhao Li and Xin Li and Yanghao Li and Yishan Li and Zhen Li and Dan Liu and Biyuan Lin and Yankai Lin and Xiang Long and Quanyu Lu and Yaxi Lu and Peiyan Luo and Hongya Lyu and Litu Ou and Yinxu Pan and Lushi Pu and Zekai Qu and Qundong Shi and Zijun Song and Jiayuan Su and Zhou Su and Ao Sun and Xianghui Sun and Peijun Tang and Fangzheng Wang and Feng Wang and Shuo Wang and Yudong Wang and Zheng Wang and Yesai Wu and Zhenyu Xiao and Jie Xie and Zihao Xie and Xiaoyue Xu and Yukun Yan and Jiarui Yuan and Jinqian Zhang and Kaihuo Zhang and Lei Zhang and Linyue Zhang and Xueren Zhang and Yudi Zhang and Hengyu Zhao and Weilin Zhao and Weilun Zhao and Yuanqian Zhao and Zhi Zheng and Chuyue Zhou and Ge Zhou and Jie Zhou and Wei Zhou and Yanghao Zhou and Zihan Zhou and Zixuan Zhou and Zhiyuan Liu and Guoyang Zeng and Chao Jia and Dahai Li and Maosong Sun},
  journal={arXiv preprint arXiv:2506.07900},
  year={2025}
}

@article{vanderweele2013three,
  title={A three-way decomposition of a total effect into direct, indirect, and interactive effects},
  author={VanderWeele, Tyler J},
  journal={Epidemiology},
  volume={24},
  number={2},
  pages={224--232},
  year={2013},
  publisher={LWW}
}

@article{marks2023geometry,
  title={The geometry of truth: Emergent linear structure in large language model representations of true/false datasets},
  author={Marks, Samuel and Tegmark, Max},
  journal={arXiv preprint arXiv:2310.06824},
  year={2023}
}

@article{arditi2024refusal,
  title={Refusal in language models is mediated by a single direction},
  author={Arditi, Andy and Obeso, Oscar and Syed, Aaquib and Paleka, Daniel and Panickssery, Nina and Gurnee, Wes and Nanda, Neel},
  journal={Advances in Neural Information Processing Systems},
  volume={37},
  pages={136037--136083},
  year={2024}
}

@article{guha2025openthoughts,
  title={Openthoughts: Data recipes for reasoning models},
  author={Etash Guha and Ryan Marten and Sedrick Keh and Negin Raoof and Georgios Smyrnis and Hritik Bansal and Marianna Nezhurina and Jean Mercat and Trung Vu and Zayne Sprague and Ashima Suvarna and Benjamin Feuer and Liangyu Chen and Zaid Khan and Eric Frankel and Sachin Grover and Caroline Choi and Niklas Muennighoff and Shiye Su and Wanjia Zhao and John Yang and Shreyas Pimpalgaonkar and Kartik Sharma and Charlie Cheng-Jie Ji and Yichuan Deng and Sarah Pratt and Vivek Ramanujan and Jon Saad-Falcon and Jeffrey Li and Achal Dave and Alon Albalak and Kushal Arora and Blake Wulfe and Chinmay Hegde and Greg Durrett and Sewoong Oh and Mohit Bansal and Saadia Gabriel and Aditya Grover and Kai-Wei Chang and Vaishaal Shankar and Aaron Gokaslan and Mike A. Merrill and Tatsunori Hashimoto and Yejin Choi and Jenia Jitsev and Reinhard Heckel and Maheswaran Sathiamoorthy and Alexandros G. Dimakis and Ludwig Schmidt},
  journal={arXiv preprint arXiv:2506.04178},
  year={2025}
}

@article{guo2025deepseek,
  title={DeepSeek-R1 incentivizes reasoning in LLMs through reinforcement learning},
  author={Guo, Daya and Yang, Dejian and Zhang, Haowei and Song, Junxiao and Wang, Peiyi and Zhu, Qihao and Xu, Runxin and Zhang, Ruoyu and Ma, Shirong and Bi, Xiao and Zhang, Xiaokang and Yu, Xingkai and Wu, Yu and Wu, Z. F. and Gou, Zhibin and Shao, Zhihong and Li, Zhuoshu and Gao, Ziyi and Liu, Aixin and Xue, Bing and Wang, Bingxuan and Wu, Bochao and Feng, Bei and Lu, Chengda and Zhao, Chenggang and Deng, Chengqi and Ruan, Chong and Dai, Damai and Chen, Deli and Ji, Dongjie and Li, Erhang and Lin, Fangyun and Dai, Fucong and Luo, Fuli and Hao, Guangbo and Chen, Guanting and Li, Guowei and Zhang, H. and Xu, Hanwei and Ding, Honghui and Gao, Huazuo and Qu, Hui and Li, Hui and Guo, Jianzhong and Li, Jiashi and Chen, Jingchang and Yuan, Jingyang and Tu, Jinhao and Qiu, Junjie and Li, Junlong and Cai, J. L. and Ni, Jiaqi and Liang, Jian and Chen, Jin and Dong, Kai and Hu, Kai and You, Kaichao and Gao, Kaige and Guan, Kang and Huang, Kexin and Yu, Kuai and Wang, Lean and Zhang, Lecong and Zhao, Liang and Wang, Litong and Zhang, Liyue and Xu, Lei and Xia, Leyi and Zhang, Mingchuan and Zhang, Minghua and Tang, Minghui and Zhou, Mingxu and Li, Meng and Wang, Miaojun and Li, Mingming and Tian, Ning and Huang, Panpan and Zhang, Peng and Wang, Qiancheng and Chen, Qinyu and Du, Qiushi and Ge, Ruiqi and Zhang, Ruisong and Pan, Ruizhe and Wang, Runji and Chen, R. J. and Jin, R. L. and Chen, Ruyi and Lu, Shanghao and Zhou, Shangyan and Chen, Shanhuang and Ye, Shengfeng and Wang, Shiyu and Yu, Shuiping and Zhou, Shunfeng and Pan, Shuting and Li, S. S. and Zhou, Shuang and Wu, Shaoqing and Yun, Tao and Pei, Tian and Sun, Tianyu and Wang, T. and Zeng, Wangding and Liu, Wen and Liang, Wenfeng and Gao, Wenjun and Yu, Wenqin and Zhang, Wentao and Xiao, W. L. and An, Wei and Liu, Xiaodong and Wang, Xiaohan and Chen, Xiaokang and Nie, Xiaotao and Cheng, Xin and Liu, Xin and Xie, Xin and Liu, Xingchao and Yang, Xinyu and Li, Xinyuan and Su, Xuecheng and Lin, Xuheng and Li, X. Q. and Jin, Xiangyue and Shen, Xiaojin and Chen, Xiaosha and Sun, Xiaowen and Wang, Xiaoxiang and Song, Xinnan and Zhou, Xinyi and Wang, Xianzu and Shan, Xinxia and Li, Y. K. and Wang, Y. Q. and Wei, Y. X. and Zhang, Yang and Xu, Yanhong and Li, Yao and Zhao, Yao and Sun, Yaofeng and Wang, Yaohui and Yu, Yi and Zhang, Yichao and Shi, Yifan and Xiong, Yiliang and He, Ying and Piao, Yishi and Wang, Yisong and Tan, Yixuan and Ma, Yiyang and Liu, Yiyuan and Guo, Yongqiang and Ou, Yuan and Wang, Yuduan and Gong, Yue and Zou, Yuheng and He, Yujia and Xiong, Yunfan and Luo, Yuxiang and You, Yuxiang and Liu, Yuxuan and Zhou, Yuyang and Zhu, Y. X. and Huang, Yanping and Li, Yaohui and Zheng, Yi and Zhu, Yuchen and Ma, Yunxian and Tang, Ying and Zha, Yukun and Yan, Yuting and Ren, Z. Z. and Ren, Zehui and Sha, Zhangli and Fu, Zhe and Xu, Zhean and Xie, Zhenda and Zhang, Zhengyan and Hao, Zhewen and Ma, Zhicheng and Yan, Zhigang and Wu, Zhiyu and Gu, Zihui and Zhu, Zijia and Liu, Zijun and Li, Zilin and Xie, Ziwei and Song, Ziyang and Pan, Zizheng and Huang, Zhen and Xu, Zhipeng and Zhang, Zhongyu and Zhang, Zhen},
  journal={Nature},
  volume={645},
  number={8081},
  pages={633--638},
  year={2025},
  publisher={Nature Publishing Group UK London}
}

@article{shao2024deepseekmath,
  title={Deepseekmath: Pushing the limits of mathematical reasoning in open language models},
  author={Zhihong Shao and Peiyi Wang and Qihao Zhu and Runxin Xu and Junxiao Song and Xiao Bi and Haowei Zhang and Mingchuan Zhang and Y. K. Li and Y. Wu and Daya Guo},
  journal={arXiv preprint arXiv:2402.03300},
  year={2024}
}

@inproceedings{
zhao2026self,
title={Self-Distilled Reasoner: On-Policy Self-Distillation for Large Language Models},
author={Siyan Zhao and Zhihui Xie and Mengchen Liu and Jing Huang and Guan Pang and Feiyu Chen and Aditya Grover},
booktitle={Forty-third International Conference on Machine Learning},
year={2026},
url={https://openreview.net/forum?id=Jpxfof0EaS}
}

@article{yu2026dapo,
  title={Dapo: An open-source llm reinforcement learning system at scale},
  author={Qiying Yu and Zheng Zhang and Ruofei Zhu and Yufeng Yuan and Xiaochen Zuo and Yu Yue and Weinan Dai and Tiantian Fan and Gaohong Liu and Lingjun Liu and Xin Liu and Haibin Lin and Zhiqi Lin and Bole Ma and Guangming Sheng and Yuxuan Tong and Chi Zhang and Mofan Zhang and Wang Zhang and Hang Zhu and Jinhua Zhu and Jiaze Chen and Jiangjie Chen and Chengyi Wang and Hongli Yu and Yuxuan Song and Xiangpeng Wei and Hao Zhou and Jingjing Liu and Wei-Ying Ma and Ya-Qin Zhang and Lin Yan and Mu Qiao and Yonghui Wu and Mingxuan Wang},
  journal={Advances in Neural Information Processing Systems},
  volume={38},
  pages={113222--113244},
  year={2026}
}

@misc{aime24,
      title={American Invitational Mathematics Examination (AIME) 2024}, 
      author={Zhang, Yifan and Math-AI, Team},
      year={2024},
}

@misc{aime25,
      title={American Invitational Mathematics Examination (AIME) 2025}, 
      author={Zhang, Yifan and Math-AI, Team},
      year={2025},
}

@inproceedings{
dekoninck2026matharena,
title={Beyond Benchmarks: MathArena as an Evaluation Platform for Mathematics with {LLM}s},
author={Jasper Dekoninck and Nikola Jovanovi{\'c} and Tim Gehrunger and K{\'a}ri R{\"o}gnvaldsson and Ivo Petrov and Chenhao Sun and Martin Vechev},
booktitle={3rd AI for Math Workshop: Toward Self-Evolving Scientific Agents},
year={2026},
url={https://openreview.net/forum?id=DmPE4byHuN}
}

@misc{Kydlicek_Math-Verify_Math_Verification,
author = {Kydlíček, Hynek},
title = {{Math-Verify: Math Verification Library}},
year={2025},
}

@inproceedings{rimsky2024steering,
  title={Steering llama 2 via contrastive activation addition},
  author={Rimsky, Nina and Gabrieli, Nick and Schulz, Julian and Tong, Meg and Hubinger, Evan and Turner, Alexander},
  booktitle={Proceedings of the 62nd Annual Meeting of the Association for Computational Linguistics (Volume 1: Long Papers)},
  pages={15504--15522},
  year={2024}
}

@misc{deepseekv32,
      title={{DeepSeek-V3.2}: Pushing the Frontier of Open Large Language Models}, 
      author={DeepSeek-AI and Aixin Liu and Aoxue Mei and Bangcai Lin and Bing Xue and Bingxuan Wang and Bingzheng Xu and Bochao Wu and Bowei Zhang and Chaofan Lin and Chen Dong and Chengda Lu and Chenggang Zhao and Chengqi Deng and Chenhao Xu and Chong Ruan and Damai Dai and Daya Guo and Dejian Yang and Deli Chen and Erhang Li and Fangqi Zhou and Fangyun Lin and Fucong Dai and Guangbo Hao and Guanting Chen and Guowei Li and H. Zhang and Hanwei Xu and Hao Li and Haofen Liang and Haoran Wei and Haowei Zhang and Haowen Luo and Haozhe Ji and Honghui Ding and Hongxuan Tang and Huanqi Cao and Huazuo Gao and Hui Qu and Hui Zeng and Jialiang Huang and Jiashi Li and Jiaxin Xu and Jiewen Hu and Jingchang Chen and Jingting Xiang and Jingyang Yuan and Jingyuan Cheng and Jinhua Zhu and Jun Ran and Junguang Jiang and Junjie Qiu and Junlong Li and Junxiao Song and Kai Dong and Kaige Gao and Kang Guan and Kexin Huang and Kexing Zhou and Kezhao Huang and Kuai Yu and Lean Wang and Lecong Zhang and Lei Wang and Liang Zhao and Liangsheng Yin and Lihua Guo and Lingxiao Luo and Linwang Ma and Litong Wang and Liyue Zhang and M. S. Di and M. Y Xu and Mingchuan Zhang and Minghua Zhang and Minghui Tang and Mingxu Zhou and Panpan Huang and Peixin Cong and Peiyi Wang and Qiancheng Wang and Qihao Zhu and Qingyang Li and Qinyu Chen and Qiushi Du and Ruiling Xu and Ruiqi Ge and Ruisong Zhang and Ruizhe Pan and Runji Wang and Runqiu Yin and Runxin Xu and Ruomeng Shen and Ruoyu Zhang and S. H. Liu and Shanghao Lu and Shangyan Zhou and Shanhuang Chen and Shaofei Cai and Shaoyuan Chen and Shengding Hu and Shengyu Liu and Shiqiang Hu and Shirong Ma and Shiyu Wang and Shuiping Yu and Shunfeng Zhou and Shuting Pan and Songyang Zhou and Tao Ni and Tao Yun and Tian Pei and Tian Ye and Tianyuan Yue and Wangding Zeng and Wen Liu and Wenfeng Liang and Wenjie Pang and Wenjing Luo and Wenjun Gao and Wentao Zhang and Xi Gao and Xiangwen Wang and Xiao Bi and Xiaodong Liu and Xiaohan Wang and Xiaokang Chen and Xiaokang Zhang and Xiaotao Nie and Xin Cheng and Xin Liu and Xin Xie and Xingchao Liu and Xingkai Yu and Xingyou Li and Xinyu Yang and Xinyuan Li and Xu Chen and Xuecheng Su and Xuehai Pan and Xuheng Lin and Xuwei Fu and Y. Q. Wang and Yang Zhang and Yanhong Xu and Yanru Ma and Yao Li and Yao Li and Yao Zhao and Yaofeng Sun and Yaohui Wang and Yi Qian and Yi Yu and Yichao Zhang and Yifan Ding and Yifan Shi and Yiliang Xiong and Ying He and Ying Zhou and Yinmin Zhong and Yishi Piao and Yisong Wang and Yixiao Chen and Yixuan Tan and Yixuan Wei and Yiyang Ma and Yiyuan Liu and Yonglun Yang and Yongqiang Guo and Yongtong Wu and Yu Wu and Yuan Cheng and Yuan Ou and Yuanfan Xu and Yuduan Wang and Yue Gong and Yuhan Wu and Yuheng Zou and Yukun Li and Yunfan Xiong and Yuxiang Luo and Yuxiang You and Yuxuan Liu and Yuyang Zhou and Z. F. Wu and Z. Z. Ren and Zehua Zhao and Zehui Ren and Zhangli Sha and Zhe Fu and Zhean Xu and Zhenda Xie and Zhengyan Zhang and Zhewen Hao and Zhibin Gou and Zhicheng Ma and Zhigang Yan and Zhihong Shao and Zhixian Huang and Zhiyu Wu and Zhuoshu Li and Zhuping Zhang and Zian Xu and Zihao Wang and Zihui Gu and Zijia Zhu and Zilin Li and Zipeng Zhang and Ziwei Xie and Ziyi Gao and Zizheng Pan and Zongqing Yao and Bei Feng and Hui Li and J. L. Cai and Jiaqi Ni and Lei Xu and Meng Li and Ning Tian and R. J. Chen and R. L. Jin and S. S. Li and Shuang Zhou and Tianyu Sun and X. Q. Li and Xiangyue Jin and Xiaojin Shen and Xiaosha Chen and Xinnan Song and Xinyi Zhou and Y. X. Zhu and Yanping Huang and Yaohui Li and Yi Zheng and Yuchen Zhu and Yunxian Ma and Zhen Huang and Zhipeng Xu and Zhongyu Zhang and Dongjie Ji and Jian Liang and Jianzhong Guo and Jin Chen and Leyi Xia and Miaojun Wang and Mingming Li and Peng Zhang and Ruyi Chen and Shangmian Sun and Shaoqing Wu and Shengfeng Ye and T. Wang and W. L. Xiao and Wei An and Xianzu Wang and Xiaowen Sun and Xiaoxiang Wang and Ying Tang and Yukun Zha and Zekai Zhang and Zhe Ju and Zhen Zhang and Zihua Qu},
      year={2025},
      eprint={2512.02556},
      archivePrefix={arXiv},
      primaryClass={cs.CL},
      url={https://arxiv.org/abs/2512.02556}, 
}

@misc{gemma4,
  title={{Gemma 4} Model Card},
  author={{Google DeepMind}},
  year={2026},
  howpublished={Google AI for Developers},
  url={https://ai.google.dev/gemma/docs/core/model_card_4}
}

@misc{nvidia2026nemotron3nanoomni,
      title={Nemotron 3 Nano Omni: Efficient and Open Multimodal Intelligence},
      author={NVIDIA},
      year={2026},
      eprint={2604.24954},
      archivePrefix={arXiv},
      primaryClass={cs.LG},
      url={https://arxiv.org/abs/2604.24954},
}

@misc{kimiteam2026kimik25visualagentic,
      title={Kimi K2.5: Visual Agentic Intelligence}, 
      author={KimiTeam and Tongtong Bai and Yifan Bai and Yiping Bao and S. H. Cai and Yuan Cao and Y. Charles and H. S. Che and Cheng Chen and Guanduo Chen and Huarong Chen and Jia Chen and Jiahao Chen and Jianlong Chen and Jun Chen and Kefan Chen and Liang Chen and Ruijue Chen and Xinhao Chen and Yanru Chen and Yanxu Chen and Yicun Chen and Yimin Chen and Yingjiang Chen and Yuankun Chen and Yujie Chen and Yutian Chen and Zhirong Chen and Ziwei Chen and Dazhi Cheng and Minghan Chu and Jialei Cui and Jiaqi Deng and Muxi Diao and Hao Ding and Mengfan Dong and Mengnan Dong and Yuxin Dong and Yuhao Dong and Angang Du and Chenzhuang Du and Dikang Du and Lingxiao Du and Yulun Du and Yu Fan and Shengjun Fang and Qiulin Feng and Yichen Feng and Garimugai Fu and Kelin Fu and Hongcheng Gao and Tong Gao and Yuyao Ge and Shangyi Geng and Chengyang Gong and Xiaochen Gong and Zhuoma Gongque and Qizheng Gu and Xinran Gu and Yicheng Gu and Longyu Guan and Yuanying Guo and Xiaoru Hao and Weiran He and Wenyang He and Yunjia He and Chao Hong and Hao Hu and Jiaxi Hu and Yangyang Hu and Zhenxing Hu and Ke Huang and Ruiyuan Huang and Weixiao Huang and Zhiqi Huang and Tao Jiang and Zhejun Jiang and Xinyi Jin and Yu Jing and Guokun Lai and Aidi Li and C. Li and Cheng Li and Fang Li and Guanghe Li and Guanyu Li and Haitao Li and Haoyang Li and Jia Li and Jingwei Li and Junxiong Li and Lincan Li and Mo Li and Weihong Li and Wentao Li and Xinhang Li and Xinhao Li and Yang Li and Yanhao Li and Yiwei Li and Yuxiao Li and Zhaowei Li and Zheming Li and Weilong Liao and Jiawei Lin and Xiaohan Lin and Zhishan Lin and Zichao Lin and Cheng Liu and Chenyu Liu and Hongzhang Liu and Liang Liu and Shaowei Liu and Shudong Liu and Shuran Liu and Tianwei Liu and Tianyu Liu and Weizhou Liu and Xiangyan Liu and Yangyang Liu and Yanming Liu and Yibo Liu and Yuanxin Liu and Yue Liu and Zhengying Liu and Zhongnuo Liu and Enzhe Lu and Haoyu Lu and Zhiyuan Lu and Junyu Luo and Tongxu Luo and Yashuo Luo and Long Ma and Yingwei Ma and Shaoguang Mao and Yuan Mei and Xin Men and Fanqing Meng and Zhiyong Meng and Yibo Miao and Minqing Ni and Kun Ouyang and Siyuan Pan and Bo Pang and Yuchao Qian and Ruoyu Qin and Zeyu Qin and Jiezhong Qiu and Bowen Qu and Zeyu Shang and Youbo Shao and Tianxiao Shen and Zhennan Shen and Juanfeng Shi and Lidong Shi and Shengyuan Shi and Feifan Song and Pengwei Song and Tianhui Song and Xiaoxi Song and Hongjin Su and Jianlin Su and Zhaochen Su and Lin Sui and Jinsong Sun and Junyao Sun and Tongyu Sun and Flood Sung and Yunpeng Tai and Chuning Tang and Heyi Tang and Xiaojuan Tang and Zhengyang Tang and Jiawen Tao and Shiyuan Teng and Chaoran Tian and Pengfei Tian and Ao Wang and Bowen Wang and Chensi Wang and Chuang Wang and Congcong Wang and Dingkun Wang and Dinglu Wang and Dongliang Wang and Feng Wang and Hailong Wang and Haiming Wang and Hengzhi Wang and Huaqing Wang and Hui Wang and Jiahao Wang and Jinhong Wang and Jiuzheng Wang and Kaixin Wang and Linian Wang and Qibin Wang and Shengjie Wang and Shuyi Wang and Si Wang and Wei Wang and Xiaochen Wang and Xinyuan Wang and Yao Wang and Yejie Wang and Yipu Wang and Yiqin Wang and Yucheng Wang and Yuzhi Wang and Zhaoji Wang and Zhaowei Wang and Zhengtao Wang and Zhexu Wang and Zihan Wang and Zizhe Wang and Chu Wei and Ming Wei and Chuan Wen and Zichen Wen and Chengjie Wu and Haoning Wu and Junyan Wu and Rucong Wu and Wenhao Wu and Yuefeng Wu and Yuhao Wu and Yuxin Wu and Zijian Wu and Chenjun Xiao and Jin Xie and Xiaotong Xie and Yuchong Xie and Yifei Xin and Bowei Xing and Boyu Xu and Jianfan Xu and Jing Xu and Jinjing Xu and L. H. Xu and Lin Xu and Suting Xu and Weixin Xu and Xinbo Xu and Xinran Xu and Yangchuan Xu and Yichang Xu and Yuemeng Xu and Zelai Xu and Ziyao Xu and Junjie Yan and Yuzi Yan and Guangyao Yang and Hao Yang and Junwei Yang and Kai Yang and Ningyuan Yang and Ruihan Yang and Xiaofei Yang and Xinlong Yang and Ying Yang and Yi Yang and Yi Yang and Zhen Yang and Zhilin Yang and Zonghan Yang and Haotian Yao and Dan Ye and Wenjie Ye and Zhuorui Ye and Bohong Yin and Chengzhen Yu and Longhui Yu and Tao Yu and Tianxiang Yu and Enming Yuan and Mengjie Yuan and Xiaokun Yuan and Yang Yue and Weihao Zeng and Dunyuan Zha and Haobing Zhan and Dehao Zhang and Hao Zhang and Jin Zhang and Puqi Zhang and Qiao Zhang and Rui Zhang and Xiaobin Zhang and Y. Zhang and Yadong Zhang and Yangkun Zhang and Yichi Zhang and Yizhi Zhang and Yongting Zhang and Yu Zhang and Yushun Zhang and Yutao Zhang and Yutong Zhang and Zheng Zhang and Chenguang Zhao and Feifan Zhao and Jinxiang Zhao and Shuai Zhao and Xiangyu Zhao and Yikai Zhao and Zijia Zhao and Huabin Zheng and Ruihan Zheng and Shaojie Zheng and Tengyang Zheng and Junfeng Zhong and Longguang Zhong and Weiming Zhong and M. Zhou and Runjie Zhou and Xinyu Zhou and Zaida Zhou and Jinguo Zhu and Liya Zhu and Xinhao Zhu and Yuxuan Zhu and Zhen Zhu and Jingze Zhuang and Weiyu Zhuang and Ying Zou and Xinxing Zu},
      year={2026},
      eprint={2602.02276},
      archivePrefix={arXiv},
      primaryClass={cs.CL},
      url={https://arxiv.org/abs/2602.02276}, 
}

@misc{zou2025representationengineeringtopdownapproach,
      title={Representation Engineering: A Top-Down Approach to AI Transparency}, 
      author={Andy Zou and Long Phan and Sarah Chen and James Campbell and Phillip Guo and Richard Ren and Alexander Pan and Xuwang Yin and Mantas Mazeika and Ann-Kathrin Dombrowski and Shashwat Goel and Nathaniel Li and Michael J. Byun and Zifan Wang and Alex Mallen and Steven Basart and Sanmi Koyejo and Dawn Song and Matt Fredrikson and J. Zico Kolter and Dan Hendrycks},
      year={2025},
      eprint={2310.01405},
      archivePrefix={arXiv},
      primaryClass={cs.LG},
      url={https://arxiv.org/abs/2310.01405}, 
}

@inproceedings{NEURIPS2024_58cbe393,
 author = {Cao, Yuanpu and Zhang, Tianrong and Cao, Bochuan and Yin, Ziyi and Lin, Lu and Ma, Fenglong and Chen, Jinghui},
 booktitle = {Advances in Neural Information Processing Systems},
 doi = {10.52202/079017-1567},
 editor = {A. Globerson and L. Mackey and D. Belgrave and A. Fan and U. Paquet and J. Tomczak and C. Zhang},
 pages = {49519--49551},
 publisher = {Curran Associates, Inc.},
 title = {Personalized Steering of Large Language Models: Versatile Steering Vectors Through Bi-directional Preference Optimization},
 url = {https://proceedings.neurips.cc/paper_files/paper/2024/file/58cbe393b4254da8966780a40d023c0b-Paper-Conference.pdf},
 volume = {37},
 year = {2024}
}

@misc{chen2025personavectorsmonitoringcontrolling,
      title={Persona Vectors: Monitoring and Controlling Character Traits in Language Models}, 
      author={Runjin Chen and Andy Arditi and Henry Sleight and Owain Evans and Jack Lindsey},
      year={2025},
      eprint={2507.21509},
      archivePrefix={arXiv},
      primaryClass={cs.CL},
      url={https://arxiv.org/abs/2507.21509}, 
}

@article{vig2020investigating,
  title={Investigating gender bias in language models using causal mediation analysis},
  author={Vig, Jesse and Gehrmann, Sebastian and Belinkov, Yonatan and Qian, Sharon and Nevo, Daniel and Singer, Yaron and Shieber, Stuart},
  journal={Advances in neural information processing systems},
  volume={33},
  pages={12388--12401},
  year={2020}
}

@article{meng2022locating,
  title={Locating and editing factual associations in gpt},
  author={Meng, Kevin and Bau, David and Andonian, Alex and Belinkov, Yonatan},
  journal={Advances in neural information processing systems},
  volume={35},
  pages={17359--17372},
  year={2022}
}

@inproceedings{
wang2023interpretability,
title={Interpretability in the Wild: a Circuit for Indirect Object Identification in {GPT}-2 Small},
author={Kevin Ro Wang and Alexandre Variengien and Arthur Conmy and Buck Shlegeris and Jacob Steinhardt},
booktitle={The Eleventh International Conference on Learning Representations },
year={2023},
url={https://openreview.net/forum?id=NpsVSN6o4ul}
}

@inproceedings{
zhang2024towards,
title={Towards Best Practices of Activation Patching in Language Models: Metrics and Methods},
author={Fred Zhang and Neel Nanda},
booktitle={The Twelfth International Conference on Learning Representations},
year={2024},
url={https://openreview.net/forum?id=Hf17y6u9BC}
}

@inproceedings{NEURIPS2025_537d5aa7,
 author = {Yue, Yang and Chen, Zhiqi and Lu, Rui and Zhao, Andrew and Wang, Zhaokai and Yue, Yang and Song, Shiji and Huang, Gao},
 booktitle = {Advances in Neural Information Processing Systems},
 editor = {D. Belgrave and C. Zhang and H. Lin and R. Pascanu and P. Koniusz and M. Ghassemi and N. Chen},
 pages = {57654--57689},
 publisher = {Curran Associates, Inc.},
 title = {Does Reinforcement Learning Really Incentivize Reasoning Capacity in LLMs Beyond the Base Model?},
 url = {https://proceedings.neurips.cc/paper_files/paper/2025/file/537d5aa768c2d534016a4d06f87bc8fb-Paper-Conference.pdf},
 volume = {38},
 year = {2025}
}

@inproceedings{
karan2026reasoning,
title={Reasoning with Sampling: Your Base Model is Smarter Than You Think},
author={Aayush Karan and Yilun Du},
booktitle={The Fourteenth International Conference on Learning Representations},
year={2026},
url={https://openreview.net/forum?id=Vsgq2ldr4K}
}

@misc{snell2024scalingllmtesttimecompute,
      title={Scaling LLM Test-Time Compute Optimally can be More Effective than Scaling Model Parameters}, 
      author={Charlie Snell and Jaehoon Lee and Kelvin Xu and Aviral Kumar},
      year={2024},
      eprint={2408.03314},
      archivePrefix={arXiv},
      primaryClass={cs.LG},
      url={https://arxiv.org/abs/2408.03314}, 
}

@article{sui2025stop,
title={Stop Overthinking: A Survey on Efficient Reasoning for Large Language Models},
author={Yang Sui and Yu-Neng Chuang and Guanchu Wang and Jiamu Zhang and Tianyi Zhang and Jiayi Yuan and Hongyi Liu and Andrew Wen and Shaochen Zhong and Na Zou and Hanjie Chen and Xia Hu},
journal={Transactions on Machine Learning Research},
issn={2835-8856},
year={2025},
url={https://openreview.net/forum?id=HvoG8SxggZ},
note={}
}

@misc{openai_reasoning,
  author       = {{OpenAI}},
  title        = {Document of Reasoning Models},
  howpublished = {\url{https://developers.openai.com/api/docs/guides/reasoning?api-mode=responses}},
  note         = {OpenAI API Documentation. Accessed: 2026-09-11},
    year={2026}
}

@misc{anthropic_effort,
  author       = {{Anthropic}},
  title        = {Document of Reasoning Effort},
  howpublished = {\url{https://platform.claude.com/docs/en/build-with-claude/effort}},
  note         = {Claude Platform Docs. Accessed: 2026-09-11},
    year={2026}
}

@inproceedings{
li2026unifying,
title={Unifying Group-Relative and Self-Distillation Policy Optimization via Sample Routing},
author={Gengsheng Li and Tianyu Yang and Junfeng Fang and Mingyang Song and Mao Zheng and Haiyun Guo and Dan Zhang and Jinqiao Wang and Tat-Seng Chua},
booktitle={Third Conference on Language Modeling},
year={2026},
url={https://openreview.net/forum?id=P2OuWwZspP}
}

@inproceedings{
hubotter2026reinforcement,
title={Reinforcement Learning via Self-Distillation},
author={Jonas H{\"u}botter and Frederike L{\"u}beck and Lejs Deen Behric and Anton Baumann and Marco Bagatella and Daniel Marta and Ido Hakimi and Idan Shenfeld and Thomas Kleine Buening and Carlos Guestrin and Andreas Krause},
booktitle={Forty-third International Conference on Machine Learning},
year={2026},
url={https://openreview.net/forum?id=QkfkxyRizZ}
}

@misc{yang2026learningteachergeneralizedonpolicy,
      title={Learning beyond Teacher: Generalized On-Policy Distillation with Reward Extrapolation}, 
      author={Wenkai Yang and Weijie Liu and Ruobing Xie and Kai Yang and Saiyong Yang and Yankai Lin},
      year={2026},
      eprint={2602.12125},
      archivePrefix={arXiv},
      primaryClass={cs.LG},
      url={https://arxiv.org/abs/2602.12125}, 
}

@inproceedings{
lin2026resrl,
title={Res{RL}: Boosting {LLM} Reasoning via Negative Sample Projection Residual Reinforcement Learning},
author={Zihan Lin and Xiaohan Wang and Jie Cao and Jiajun Chai and Li Wang and Xiaodong Lu and Wei Lin and Ran He and Guojun Yin},
booktitle={Forty-third International Conference on Machine Learning},
year={2026},
url={https://openreview.net/forum?id=kmN9ozKtGh}
}

@misc{huang2026bootstrappingexplorationgrouplevelnatural,
      title={Bootstrapping Exploration with Group-Level Natural Language Feedback in Reinforcement Learning}, 
      author={Lei Huang and Xiang Cheng and Chenxiao Zhao and Guobin Shen and Junjie Yang and Xiaocheng Feng and Yuxuan Gu and Xing Yu and Bing Qin},
      year={2026},
      eprint={2603.04597},
      archivePrefix={arXiv},
      primaryClass={cs.CL},
      url={https://arxiv.org/abs/2603.04597}, 
}

@inproceedings{
zhu2025the,
title={The Surprising Effectiveness of Negative Reinforcement in {LLM} Reasoning},
author={Xinyu Zhu and Mengzhou Xia and Zhepei Wei and Wei-Lin Chen and Danqi Chen and Yu Meng},
booktitle={The Thirty-ninth Annual Conference on Neural Information Processing Systems},
year={2025},
url={https://openreview.net/forum?id=ftVlLG9cks}
}

@misc{li2026onpolicyselfdistillationsupervision,
      title={On-Policy Self-Distillation without Any Supervision}, 
      author={Yijiang Li and Bingyang Wang and Yijun Liang and Yunjie Tian and Di Fu and Nuno Vasconcelos},
      year={2026},
      eprint={2608.06296},
      archivePrefix={arXiv},
      primaryClass={cs.LG},
      url={https://arxiv.org/abs/2608.06296}, 
}

@misc{xu2026agpoasymmetricgrouppolicy,
      title={AGPO: Asymmetric Group Policy Optimization for Verifiable Reasoning and Search Ads Relevance at JD}, 
      author={Yang Xu and Kun Yao and Yiming Deng and Zheng Fang and Kai Ming Ting and Ming Pang},
      year={2026},
      eprint={2605.05826},
      archivePrefix={arXiv},
      primaryClass={cs.AI},
      url={https://arxiv.org/abs/2605.05826}, 
}

@misc{ding2026doesonpolicydistillationreally,
      title={Does On-Policy Distillation Really Distill? From Noisy Teacher to Self-Improvement}, 
      author={Yi Ding and Ruqi Zhang},
      year={2026},
      eprint={2608.31046},
      archivePrefix={arXiv},
      primaryClass={cs.LG},
      url={https://arxiv.org/abs/2608.31046}, 
}

@inproceedings{NEURIPS2025_b8026fe0,
 author = {Zhu, Xinyu and Xia, Mengzhou and Wei, Zhepei and Chen, Wei-Lin and Chen, Danqi and Meng, Yu},
 booktitle = {Advances in Neural Information Processing Systems},
 doi = {10.52202/085713-4220},
 editor = {D. Belgrave and C. Zhang and H. Lin and R. Pascanu and P. Koniusz and M. Ghassemi and N. Chen},
 pages = {126546--126573},
 publisher = {Curran Associates, Inc.},
 title = {The Surprising Effectiveness of Negative Reinforcement in LLM Reasoning},
 url = {https://proceedings.neurips.cc/paper_files/paper/2025/file/b8026fe01f5f8f7a80a1a9bf01d08cad-Paper-Conference.pdf},
 volume = {38, Main Conference},
 year = {2025}
}

@inproceedings{gan-etal-2026-thinking,
    title = "Thinking-Based Non-Thinking: Solving the Reward Hacking Problem in Training Hybrid Reasoning Models via Reinforcement Learning",
    author = "Gan, Siyuan  and
      Liu, Jiaheng  and
      Wang, Boyan  and
      Yang, Tianpei  and
      Miao, Runqing  and
      Zhang, Yuyao  and
      Meng, Fanyu  and
      Feng, Junlan  and
      Meng, Linjian  and
      Huo, Jing  and
      Gao, Yang",
    editor = "Liakata, Maria  and
      Moreira, Viviane P.  and
      Zhang, Jiajun  and
      Jurgens, David",
    booktitle = "Proceedings of the 64th Annual Meeting of the {A}ssociation for {C}omputational {L}inguistics (Volume 1: Long Papers)",
    month = jul,
    year = "2026",
    address = "San Diego, California, United States",
    publisher = "Association for Computational Linguistics",
    url = "https://aclanthology.org/2026.acl-long.2122/",
    doi = "10.18653/v1/2026.acl-long.2122",
    pages = "45754--45771",
    ISBN = "979-8-89176-390-6",
}

@inproceedings{zhang-etal-2025-adaptthink,
    title = "{A}dapt{T}hink: Reasoning Models Can Learn When to Think",
    author = "Zhang, Jiajie  and
      Lin, Nianyi  and
      Hou, Lei  and
      Feng, Ling  and
      Li, Juanzi",
    editor = "Christodoulopoulos, Christos  and
      Chakraborty, Tanmoy  and
      Rose, Carolyn  and
      Peng, Violet",
    booktitle = "Proceedings of the 2025 Conference on Empirical Methods in Natural Language Processing",
    month = nov,
    year = "2025",
    address = "Suzhou, China",
    publisher = "Association for Computational Linguistics",
    url = "https://aclanthology.org/2025.emnlp-main.184/",
    doi = "10.18653/v1/2025.emnlp-main.184",
    pages = "3716--3730",
    ISBN = "979-8-89176-332-6",
}

@inproceedings{NEURIPS2025_16371a9d,
 author = {Tu, Songjun and Lin, Jiahao and Zhang, Qichao and Tian, Xiangyu and Li, Linjing and Lan, Xiangyuan and Zhao, Dongbin},
 booktitle = {Advances in Neural Information Processing Systems},
 doi = {10.52202/085713-0510},
 editor = {D. Belgrave and C. Zhang and H. Lin and R. Pascanu and P. Koniusz and M. Ghassemi and N. Chen},
 pages = {15181--15207},
 publisher = {Curran Associates, Inc.},
 title = {Learning When to Think: Shaping Adaptive Reasoning in R1-Style Models via Multi-Stage RL},
 url = {https://proceedings.neurips.cc/paper_files/paper/2025/file/16371a9d5fed65d6d78ca3a7fa6e598c-Paper-Conference.pdf},
 volume = {38, Main Conference},
 year = {2025}
}

@inproceedings{NEURIPS2025_70d46c12,
 author = {Lin, Zhengkai and Fu, Zhihang and Chen, Ze and Chen, Chao and Xie, Liang and Wang, Wenxiao and Cai, Deng and Wang, Zheng and Ye, Jieping},
 booktitle = {Advances in Neural Information Processing Systems},
 doi = {10.52202/085713-2623},
 editor = {D. Belgrave and C. Zhang and H. Lin and R. Pascanu and P. Koniusz and M. Ghassemi and N. Chen},
 pages = {78300--78347},
 publisher = {Curran Associates, Inc.},
 title = {Controlling Thinking Speed in Reasoning Models},
 url = {https://proceedings.neurips.cc/paper_files/paper/2025/file/70d46c12643e8f056a693ed43472dc35-Paper-Conference.pdf},
 volume = {38, Main Conference},
 year = {2025}
}

@misc{venhoff2025understandingreasoningthinkinglanguage,
      title={Understanding Reasoning in Thinking Language Models via Steering Vectors}, 
      author={Constantin Venhoff and Iván Arcuschin and Philip Torr and Arthur Conmy and Neel Nanda},
      year={2025},
      eprint={2506.18167},
      archivePrefix={arXiv},
      primaryClass={cs.LG},
      url={https://arxiv.org/abs/2506.18167}, 
}

@inproceedings{sinii-etal-2025-steering,
    title = "Steering {LLM} Reasoning Through Bias-Only Adaptation",
    author = "Sinii, Viacheslav  and
      Gorbatovski, Alexey  and
      Cherepanov, Artem  and
      Shaposhnikov, Boris  and
      Balagansky, Nikita  and
      Gavrilov, Daniil",
    editor = "Christodoulopoulos, Christos  and
      Chakraborty, Tanmoy  and
      Rose, Carolyn  and
      Peng, Violet",
    booktitle = "Proceedings of the 2025 Conference on Empirical Methods in Natural Language Processing",
    month = nov,
    year = "2025",
    address = "Suzhou, China",
    publisher = "Association for Computational Linguistics",
    url = "https://aclanthology.org/2025.emnlp-main.467/",
    doi = "10.18653/v1/2025.emnlp-main.467",
    pages = "9202--9211",
    ISBN = "979-8-89176-332-6"
}

@article{sheng2024hybridflow,
  title   = {Hybrid{F}low: A Flexible and Efficient RLHF Framework},
  author  = {Guangming Sheng and Chi Zhang and Zilingfeng Ye and Xibin Wu and Wang Zhang and Ru Zhang and Yanghua Peng and Haibin Lin and Chuan Wu},
  year    = {2024},
  journal = {arXiv preprint arXiv: 2409.19256}
}

@article{xu2025easysteer,
  title={Easy{S}teer: A Unified Framework for High-Performance and Extensible LLM Steering},
  author={Xu, Haolei and Mei, Xinyu and Yan, Yuchen and Zhou, Rui and Zhang, Wenqi and Lu, Weiming and Zhuang, Yueting and Shen, Yongliang},
  journal={arXiv preprint arXiv:2509.25175},
  year={2025}
}

@InProceedings{pmlr-v267-wang25i,
  title = 	 {{T}ruth{F}low: Truthful {LLM} Generation via Representation Flow Correction},
  author =       {Wang, Hanyu and Cao, Bochuan and Cao, Yuanpu and Chen, Jinghui},
  booktitle = 	 {Proceedings of the 42nd International Conference on Machine Learning},
  pages = 	 {62423--62444},
  year = 	 {2025},
  editor = 	 {Singh, Aarti and Fazel, Maryam and Hsu, Daniel and Lacoste-Julien, Simon and Berkenkamp, Felix and Maharaj, Tegan and Wagstaff, Kiri and Zhu, Jerry},
  volume = 	 {267},
  series = 	 {Proceedings of Machine Learning Research},
  month = 	 {13--19 Jul},
  publisher =    {PMLR},
  url = 	 {https://proceedings.mlr.press/v267/wang25i.html}
}
\bibliographystyle{iclr2027_conference}
\newpage
\appendix
\renewcommand{\contentsname}{Appendix Contents}
\addtocontents{toc}{\protect\setcounter{tocdepth}{2}}
\setcounter{tocdepth}{2}
\tableofcontents
\newpage
\renewcommand{\theHsection}{A\arabic{section}}

\section{Extended Related Work}
\label{app:related}

\paragraph{Prior observations of thinking in \nothink{}.}
Work on post-training the \nothink{} mode has noted that the model often
continues to think. \citet{gan-etal-2026-thinking} call this reward hacking: a
response is labeled \nothink{} because it begins with \texttt{</think>}, but
continues with exploratory reasoning and collects the higher \nothink{} reward.
They detect it from the text, by token usage and by the density of reasoning
markers, the same behavioral signal we use in \cref{sec:correlational}.
\citet{zhang-etal-2025-adaptthink} cap \nothink{} responses at a fixed token
budget to suppress it, and \citet{gan-etal-2026-thinking} make the cap per-query;
\citet{NEURIPS2025_16371a9d}, which imposes no such constraint, produces
\nothink{} responses nearly as long as its \think{} ones.
\citet{NEURIPS2025_b8026fe0} observe a related effect on Qwen3-4B in \nothink{}
mode and read it the other way, as post-training unlocking latent reasoning
ability. In a hybrid model that ability is already accessible through \think{},
so unlocking it in \nothink{} is not by itself an improvement in \nothink{}
capability.

In these works leakage is incidental---an obstacle to remove, or a side benefit.
We study it as the object of interest, and measure it in the model's internal
representations as well as in its text.

\section{Extended Discussion}
\label{app:extended-discussion}
\subsection{Why does thinking leakage deserve attention?}
\label{app:extended-discussion1}
Post-training can elicit latent abilities from the base
model~\citep{NEURIPS2025_537d5aa7, karan2026reasoning, NEURIPS2025_b8026fe0}.
Making previously inaccessible reasoning available can itself
constitute a meaningful improvement, even without creating
new abilities.
Hybrid reasoning models present a different setting:
the base model's reasoning behavior is already directly
accessible through \think{}, where thinking leakage can inflate the apparent
contribution of post-training by recruiting already accessible
\think{} behavior into \nothink{}.
A method's performance advantage may therefore reflect its greater effectiveness
at re-invoking existing reasoning rather than the intended improvement within
\nothink{}. The accuracy gain is real, but attributing it entirely to the latter
overstates what the method achieved.

Thinking leakage does not preclude useful gains in reasoning
efficiency.
For example, our GRPO experiments use a generation budget
of $8192$ tokens, forcing the model to reach a correct final
answer within a constrained reasoning budget.
Post-training may thus make existing reasoning more concise
even when its accuracy gains depend on leakage.
For research aimed at raising the model's performance ceiling,
however, such gains alone do not establish an advance beyond
the reasoning already accessible through the base model's
\think{} mode.
This requires comparison with base \think{} across inference
budgets to distinguish more efficient use of existing reasoning
from a higher attainable level of performance.

\subsection{Beyond linear response estimation.}
\label{app:extended-discussion2}
Our causal decomposition does not require $Y(\theta,\pi)$
to be linear in either model weights or thinking engagement.
The four outcomes in \cref{eq:four-outcomes} evaluate this
function for the base model and a checkpoint, with and
without steering or counter-steering.
More densely sampled checkpoints would trace how this response
changes along the training trajectory, while finer steering
sweeps over a wider range would reveal its dependence on
engagement.
Together, these measurements could support a nonlinear
response surface $Y(\theta,\pi)$, capturing finer-grained information.
The same four-outcome accounting would apply to comparisons
between selected points on this surface.

\section{Experiment Setup}
\label{app:setup}
\paragraph{Post-training methods.}
To test whether thinking leakage is specific to a particular training algorithm, we
audit three post-training methods applied in \nothink{}:
\begin{itemize}
  \item \textbf{GRPO}~\citep{guo2025deepseek, shao2024deepseekmath}: Group Relative Policy Optimization method, for which we adopt the modified variant and configuration of \citet{hubotter2026reinforcement}, trained on DAPO-Math~\citep{yu2026dapo};
  \item \textbf{SFT}: supervised fine-tuning on the reasoning traces of OpenThoughts-Math~\citep{guha2025openthoughts}, following the configuration of \citet{zhao2026self};
  \item \textbf{OPSD}: on-policy self-distillation with or without LoRA~\citep{zhao2026self}, using its original \nothink{} training recipe on OpenThoughts-Math~\citep{guha2025openthoughts}. The LoRA configuration follows the original paper's setup while full parameter fine-tuning is also evaluated.
\end{itemize}
The detailed hyperparameter setting is given in \cref{tab:hyperparameters}.

\begin{table*}[h]
\small
\centering
\setlength{\tabcolsep}{4pt}
\newcolumntype{C}{>{\centering\arraybackslash}X}
\caption{Training and evaluation hyperparameters for the three fine-tuning
recipes. Dashes denote parameters that do not apply to a given method.}
\begin{tabularx}{\textwidth}{lCCC}
    \toprule
    \textbf{Parameter} & \textbf{GRPO} & \textbf{SFT} & \textbf{OPSD} \\
    \midrule
    \textbf{General} & & & \\
    Backbones & \multicolumn{3}{c}{Qwen3-8B, Qwen3-4B, MiniCPM4.1-8B} \\
    Training data & DAPO-Math & OpenThoughts-Math& OpenThoughts-Math \\
    \midrule
    \textbf{Data} & & & \\
    Max.\ prompt length & 2048 & 2048 & 2048 \\
    Max.\ response length & 8192 & 16000 & 1024 \\
    \midrule
    \textbf{Optimization} & & & \\
    Optimizer & AdamW & AdamW & AdamW \\
    Learning rate & $1{\times}10^{-6}$ & $1{\times}10^{-5}$ & $5{\times}10^{-6}$ \\
    Warmup steps & 10 & 0 & 0 \\
    Weight decay & 0.01 & 0 & 0 \\
    Grad.\ clip norm & 1.0 & 1.0 & 0.1 \\
    Effective batch size & 32 & 32 & 32 \\
    Total steps & 500 & 100 & 100 \\
    Evaluation frequency (/step) & 100 & 25 & 25 \\
    \midrule
    \textbf{Method-specific} & & & \\
    Rollout number & 8 & -- & 1 \\
    Rollout temperature & 1.0 & -- & 1.1 \\
    Normalize advantages & True & -- & -- \\
    $\epsilon$-high & 0.28 & -- & -- \\
    KL coefficient & 0 & -- & -- \\
    Per-token KL clip (threshold) & -- & -- & $1{\times}10^{-7}$\\
    Teacher & -- & -- & frozen $\theta_0$ \\
    Distillation divergence & -- & -- & forward KL \\
    LoRA rank / $\alpha$ & -- & -- & 64 / 128 (if applied)\\
    \midrule
    \multicolumn{4}{l}{\textbf{Evaluation}} \\
    Benchmarks & \multicolumn{3}{c}{AIME24, AIME25, HMMT-Feb-25, HMMT-Nov-25} \\
    Inference engine & \multicolumn{3}{c}{vLLM} \\
    Number of rollouts & \multicolumn{3}{c}{16} \\
    Temperature & \multicolumn{3}{c}{0.6} \\
    Top-$p$ & \multicolumn{3}{c}{0.95} \\
    Max.\ generation length & \multicolumn{3}{c}{32768 (63488 for MiniCPM4.1-8B)} \\
    \bottomrule
\end{tabularx}
\label{tab:hyperparameters}
\end{table*}

\paragraph{Evaluation data and protocol.}
We evaluate on $120$ competition-math problems drawn equally from four benchmarks ($30$ each): \textbf{AIME24}~\citep{aime24}, \textbf{AIME25}~\citep{aime25}, \textbf{HMMT-Feb-2025}, and \textbf{HMMT-Nov-2025}~\citep{dekoninck2026matharena}.
For each problem we draw $16$ samples (temperature $0.6$, top-$p=0.95$) and report the mean per-sample solve rate (\emph{avg@16}). The per-sample binary score is the outcome $Y\in\{0,1\}$ of \cref{subsec:causal}, and avg@16 is its empirical expectation $Y_{a,a'}=\mathbb{E}[Y]$.
Full training and evaluation hyperparameters are listed in \cref{tab:hyperparameters}.

\paragraph{Chunked generation and the loop filter.}
Rollouts vary widely in length, and in batched generation a batch is only as fast as its longest sequence, so generating everything one-shot at the maximum budget spends most of its time on a handful of stragglers. We therefore generate in chunks of $8192$ tokens: after each round, only rollouts that reached that round's cap are continued, by re-feeding the prompt and partial response, until they terminate or the budget is reached. Before each extension we re-test the accumulated text for degenerate repetition (word-level $12$-gram repetition ratio above $0.3$) and drop any rollout that fails from subsequent rounds. 

\paragraph{Generation budgets.}
Both Qwen models use $32768$ tokens in every setting. MiniCPM4.1-8B's \think{} responses are far longer, so we raise both its base \think{} run and its steering sweep---which induces long rollouts---to its context limit $63488$ tokens, and keep $32768$ elsewhere.

\paragraph{Correctness criterion.}
Each sample is graded binary and averaged over its $16$ samples (avg@16). We extract the last \texttt{\textbackslash boxed\{\}} expression as the final answer and compare it to the gold answer by exact string match, falling back to symbolic/ numeric equivalence via \texttt{math\_verify}~\citep{Kydlicek_Math-Verify_Math_Verification}. Responses with no parseable answer, including truncated or repetition-looped generations, are counted as incorrect.


\paragraph{Direction estimation.}
All directions $\hat{v}_{\text{leak}}$ and $\hat{v}_{\text{ref}}$ (\cref{subsec:directions} and \cref{app:vref}) are estimated on the \emph{base} model $\theta_0$. The estimation set comprises $120$ problems $\times\,16$ rollouts. 

\paragraph{Style tokens.} The marker set is the union of two per-model lists, each collected from the vocabulary that model actually uses inside \texttt{<think>}. 
We refer to the two lists as $M_{\mathrm{Qwen}}$ and $M_{\mathrm{MiniCPM}}$ below. Matches are counted case-insensitively on word boundaries.
\begin{itemize}
    \item $M_{\mathrm{Qwen}}$ (13): \texttt{wait}, \texttt{hmm}, \texttt{perhaps},
    \texttt{maybe}, \texttt{actually}, \texttt{alternatively}, \texttt{seems},
    \texttt{might}, \texttt{likely}, \texttt{check}, \texttt{reconsider},
    \texttt{backtrack}, \texttt{instead}.
    \item $M_{\mathrm{MiniCPM}}$ (11): \texttt{since}, \texttt{similarly},
    \texttt{moreover}, \texttt{notice}, \texttt{note}, \texttt{another}, \texttt{suppose},
    \texttt{hence}, \texttt{recall}, \texttt{earlier}, \texttt{no}.
\end{itemize}
Taking the union matters because the two families are close to disjoint in practice --- in \think{} mode Qwen3 emits $M_{\mathrm{Qwen}}$ at $19$ per $1000$ words against $5$ for MiniCPM4.1-8B,
while MiniCPM4.1-8B emits $M_{\mathrm{MiniCPM}}$ at $19$ against $12$ --- so scoring either model with
the other's list alone would miss most of its markers. 

\paragraph{Robustness of $\hat v_{\mathrm{leak}}$}
To assess the robustness of $\hat v_{\mathrm{leak}}$, we re-estimate it on a \emph{disjoint} set of $120$ problems from the GRPO training data ($\times 1$ rollout, $120$ samples); the resulting direction remains closely aligned with the evaluation-set estimate at layer $20$ (cosine $0.945$), confirming that it is not an artifact of the evaluation distribution. It is equally insensitive to the sampling budget: subsampling a single rollout per evaluation problem recovers the $\times 16$ direction with cosine $0.998$ across all $16$ such draws. Together, these show that $\hat v_{\mathrm{leak}}$ is stable both across data distributions and across sampling budgets.

\paragraph{Interventions.}
Our causal experiments are performed by manipulating $\hat{v}_{\text{leak}}$ during \nothink{} generation---adding it in the base model (\cref{sec:sufficiency}) and subtract it in the post-trained model (\cref{sec:necessity}); implementation details are deferred to those sections.

\paragraph{Implementation.}
GRPO and SFT training were run with \texttt{verl}~\citep{sheng2024hybridflow},OPSD with the authors' public repository~\citep{zhao2026self}, and the steering and counter-steering interventions with \texttt{EasySteer}~\citep{xu2025easysteer}. All experiments were run on two NVIDIA H200 GPUs.

\section{Supplementary Analysis for
  \texorpdfstring{\cref{sec:correlational}}{Drift Detection in NoThink}}
\label{app:leakage}

This appendix supports \cref{sec:correlational} in four parts: a reference axis that tests whether the drift is specific to $\hat v_{\mathrm{leak}}$ (\cref{app:vref}), the same measurements repeated at three probe depths (\cref{app:vref-depth}), the quantities and construction behind the mode-gap figure (\cref{app:geometry}), and a breakdown of how the three post-training methods differ across the two model families \cref{app:methods}).

\subsection{Control experiment: a reference axis}
\label{app:vref}

To test whether the drift is specific to $\hat v^{(\ell)}_{\text{leak}}$, we repeat the
measurement on a second axis built from the same rollouts.

A \think{} rollout has two parts: the exploratory reasoning inside
\texttt{<think>}\,$\ldots$\,\texttt{</think>}, and the clean solution written after it. The leakage axis $\hat v^{(\ell)}_{\text{leak}}$ of \cref{eq:vleak} contrasts \nothink{}
against the rollout as a whole; contrasting it against the \think{} solution region of $\mathcal{S}(x)$ alone gives
\begin{equation}
v^{(\ell)}_{\text{ref}}
= \bar h^{(\ell)}_{\mathcal{S}}(\theta_0) - \bar h^{(\ell)}_{\mathcal{N}}(\theta_0),
\qquad
\hat v^{(\ell)}_{\text{ref}} = v^{(\ell)}_{\text{ref}} \big/
\lVert v^{(\ell)}_{\text{ref}}\rVert .
\label{eq:vref}
\end{equation} 
The two axes separate what a checkpoint may be moving toward: drift along $\hat v^{(\ell)}_{\text{leak}}$ indicates exploratory reasoning leaking into \nothink{}, while drift along $\hat v^{(\ell)}_{\text{ref}}$ would indicate a shift toward the clean solution. A method that genuinely improved direct answering, without re-invoking deliberation, should only move along the latter. 

\begin{figure*}[h]
    \centering
    \includegraphics[width=\textwidth]{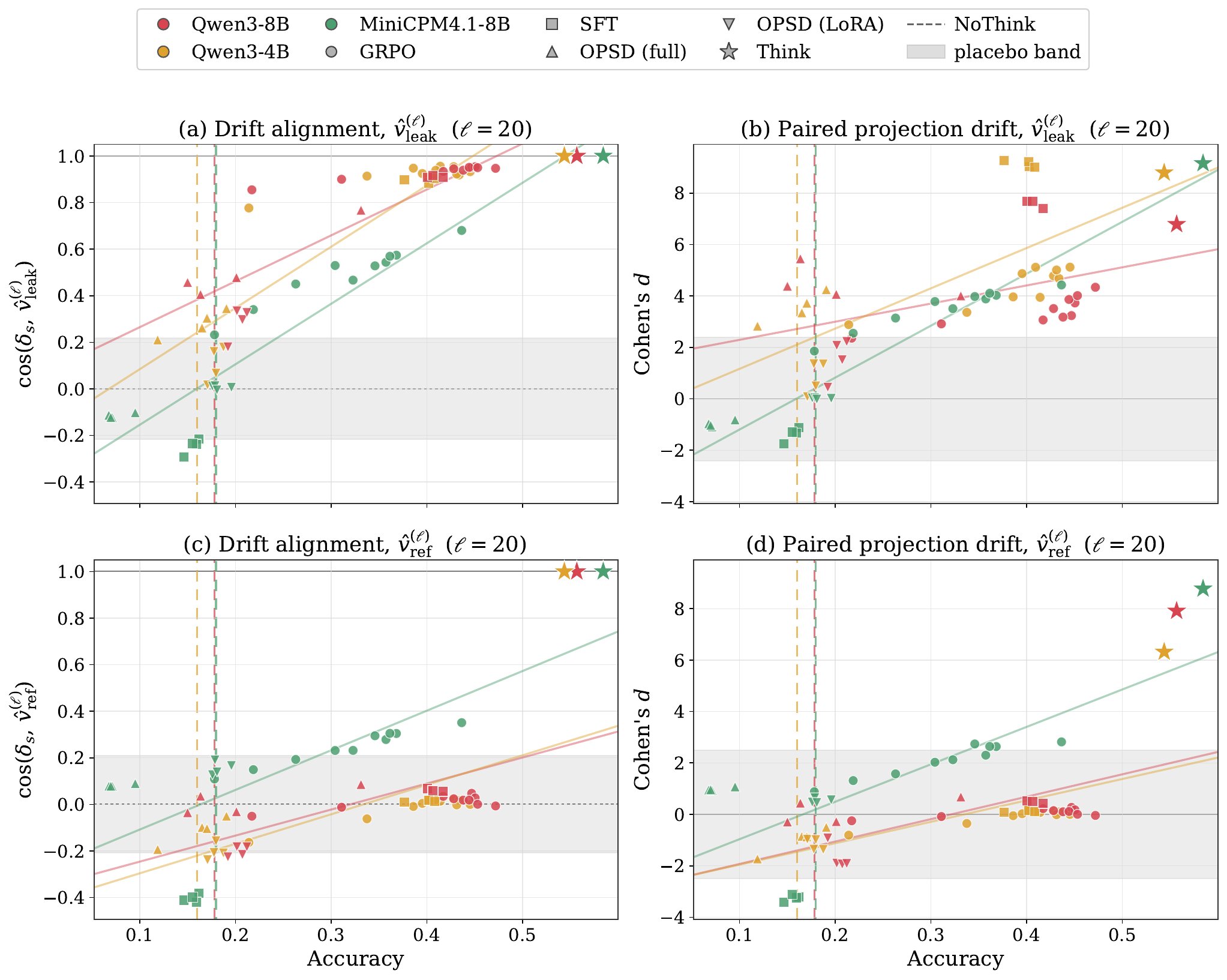}
    \caption{%
    \textbf{The drift is specific to $\hat v^{(\ell)}_{\text{leak}}$ (layer $\ell=20$).}
    Each point is one checkpoint; marker shape denotes the training method, colour the base model. $\star$ is the base model in \think{}.
    \textbf{(a), (c)} Drift alignment on $\hat{v}_{\mathrm{leak}}$ and $\hat v^{(\ell)}_{\text{ref}}$ respectively.
    \textbf{(b), (d)} Cohen's $d$ of the same drift along each axis.
    The gray \textbf{placebo band} marks the null in each panel: in (a) and (c), the alignment obtained by splitting the base model's own \nothink{} rollouts in half, so that the drift carries sampling noise only; in (b) and (d), Cohen's $d$ recomputed for the same checkpoint pair along a random unit direction. Both are drawn at the 90th percentile of the null's absolute value.
}
    \label{fig:leakage-ref}
\end{figure*}

\paragraph{Result.}
The two axes behave differently in every respect (\cref{fig:leakage-ref}).
The drift alignment points squarely at $\hat v_{\mathrm{leak}}$ (mean $\cos=0.51$, $77\%$ of checkpoints outside the null) and is unrelated to $\hat v_{\text{ref}}$ ($0.00$, $23\%$). The paired drift splits the same way: along $\hat v_{\mathrm{leak}}$ it averages $d=3.15\pm2.71$ with $66\%$ of checkpoints clearing the placebo band, while along $\hat v_{\text{ref}}$ it is centred at zero ($0.02\pm1.37$) with only $12\%$ clearing it.
Accuracy separates them as well ($\rho=+0.92$ against $+0.25$ for drift alignment, and $\rho=+0.68$ against $+0.23$ for paired drift).

Some drift along $\hat v_{\text{ref}}$ is expected---its anchor, the solution region, is part of the very rollout that anchors $\hat v_{\mathrm{leak}}$, so the two cannot be fully independent---which makes the order-of-magnitude gap the conservative reading.
Post-training does not simply sharpen the model's ability to produce a clean solution; it re-invokes the \think{} mode itself.

\begin{figure*}[h]
    \centering
    \includegraphics[width=\textwidth]{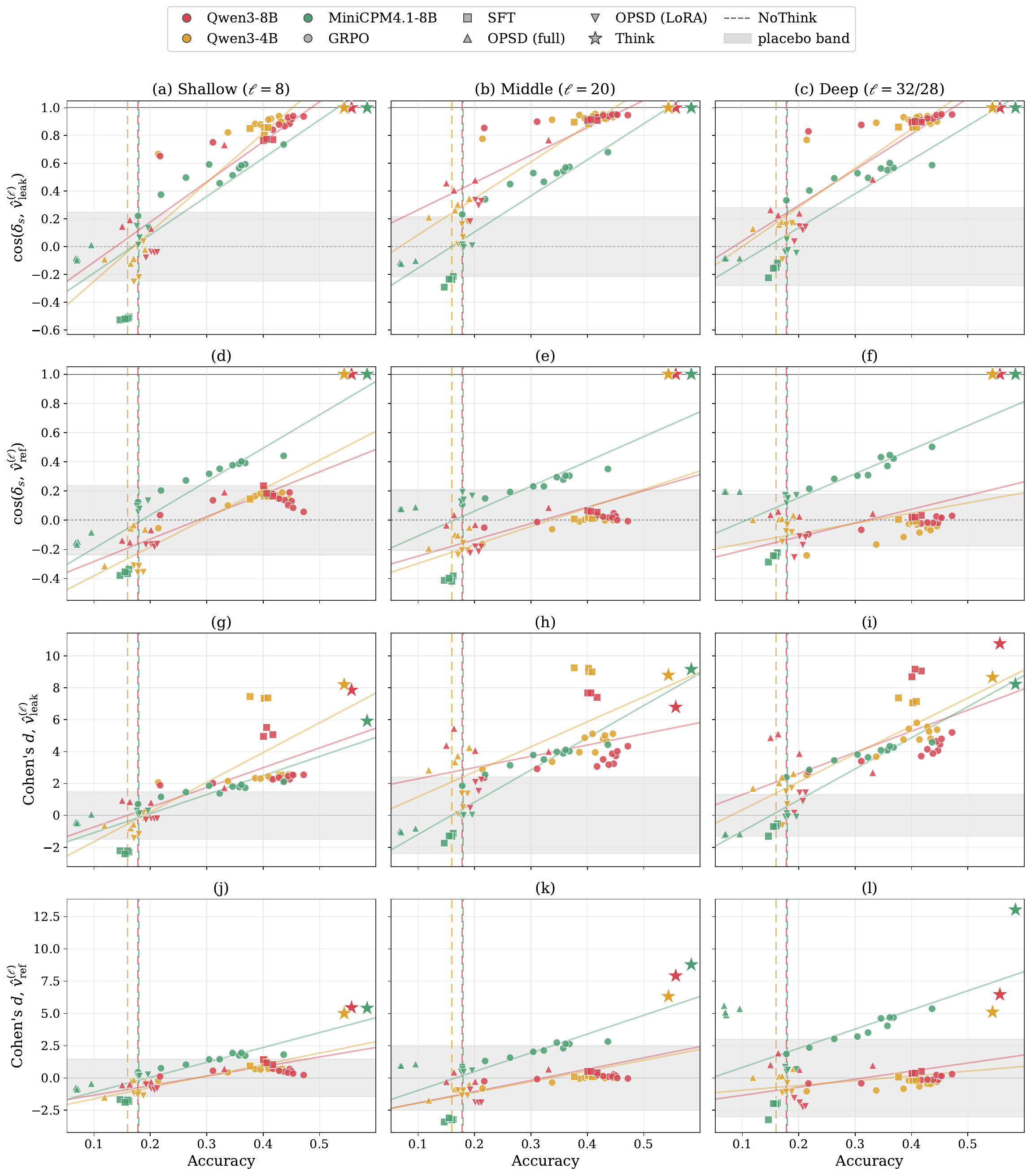}
    \caption{%
      \textbf{The leakage happens across depth.}  Rows are (measure, axis), columns are depth. Each point is one checkpoint; marker shape denotes the training method, colour the base model. The legend is the same as in~\cref{{fig:leakage-ref}}.
}
    \label{fig:leakage-depth}
\end{figure*}

\subsection{The leakage and reference axes across depth}
\label{app:vref-depth}

To test whether the drift is universal to different layers, we repeat both measures on both axes at three probe depths: shallow $\ell=8$ and middle $\ell=20$ for all models, and a deep layer set to $\ell=32$ for the $36$-layer Qwen3 models and $\ell=28$ for the $32$-layer MiniCPM4.1-8B.

\paragraph{Result.}
\Cref{fig:leakage-depth} reproduces the separation of \cref{app:vref} at every probe depth. Along $\hat v^{(\ell)}_{\text{leak}}$, $58$--$77\%$ of checkpoints clear the placebo band on either measure, and drift alignment tracks \nothink{} accuracy for every model at every layer ($\rho=0.83$--$0.93$); along $\hat v^{(\ell)}_{\text{ref}}$ only $12$--$29\%$ clear it.

Depth scales the leakage effect without changing it: the paired drift roughly doubles from
the shallow to the middle probe ($d=1.49$ to $3.15$) and is then unchanged at the deep layer ($3.17$). On the $\hat v^{(\ell)}_{\text{ref}}$, it stays near zero for the Qwen models at every depth ($|d|\le0.41$) and it grows with depth for the MiniCPM4.1-8B ($0.23$, $0.64$, $2.35$), where the two regions are least separable.

\begin{table}[h]
  \centering
  \caption{Fidelity of \cref{fig:vleak-geometry-a}. \emph{Drawn} is measured from the
  final two-dimensional coordinates; \emph{true} is recomputed in $\mathbb{R}^{4096}$
  from the raw activations. \emph{Abs.\ err.} means absolute error and \emph{Rel.\ err.} is the corresponding relative error.}
  \label{tab:fig-a-fidelity}
  \begin{tabular}{llrrrr}
    \toprule
    Quantity & Checkpoint & True & Drawn & Abs.\ err. & Rel.\ err. \\
    \midrule
    $\|v_{\mathrm{leak}}\|$ & base      & $15.426$ & $15.426$ & $-0.000$ & $-0.00\%$ \\
                            & GRPO@200  & $8.810$  & $8.777$  & $-0.034$ & $-0.38\%$ \\
                            & GRPO@500  & $7.178$  & $7.142$  & $-0.037$ & $-0.51\%$ \\
    \addlinespace \midrule
    rotation vs.\ base      & GRPO@200  & $20.270\degree$ & $19.555\degree$ & $-0.715\degree$ & $-3.53\%$ \\
                            & GRPO@500  & $26.354\degree$ & $25.844\degree$ & $-0.510\degree$ & $-1.94\%$ \\
    \addlinespace \midrule
    $\|\bar h_{\mathcal T}\|$ & base    & $72.318$ & $73.525$ & $+1.207$ & $+1.67\%$ \\
                            & GRPO@200  & $72.149$ & $73.525$ & $+1.376$ & $+1.91\%$ \\
                            & GRPO@500  & $71.927$ & $73.525$ & $+1.598$ & $+2.22\%$ \\
    \addlinespace \midrule
    $\|\bar h_{\mathcal N}\|$ & base    & $76.975$ & $78.515$ & $+1.540$ & $+2.00\%$ \\
                            & GRPO@200  & $74.429$ & $73.087$ & $-1.342$ & $-1.80\%$ \\
                            & GRPO@500  & $73.700$ & $72.307$ & $-1.393$ & $-1.89\%$ \\
    \addlinespace \midrule
    $\phi$                  & base      & $11.312\degree$ & $11.024\degree$ & $-0.288\degree$ & $-2.54\%$ \\
                            & GRPO@200  & $6.658\degree$  & $6.855\degree$  & $+0.198\degree$ & $+2.97\%$ \\
                            & GRPO@500  & $5.476\degree$  & $5.532\degree$  & $+0.056\degree$ & $+1.02\%$ \\
    \bottomrule
  \end{tabular}
\end{table}

\subsection{Mode-gap geometry: quantities, construction, and fidelity}
\label{app:geometry}

\subsubsection{Measured quantities}
Following \cref{eq:hbar}, we compute the two mode centroids of each checkpoint at $\ell=20$ in the model's native $d=4096$ dimensions. We drop the layer and checkpoint indices below, writing $\bar h_{\mathcal T},\bar h_{\mathcal N}$ for the centroids, $v_{\mathrm{leak}}=\bar h_{\mathcal T}-\bar h_{\mathcal N}$ for their difference, and $\phi$ for the angle between them. Their values are listed in the \emph{true} column of
\cref{tab:fig-a-fidelity}.

\paragraph{Decomposing the change in $\|v_{\mathrm{leak}}\|$.}
Writing $\|v_{\mathrm{leak}}\|^2=\|\bar h_{\mathcal T}\|^2+\|\bar h_{\mathcal N}\|^2-2\|\bar h_{\mathcal T}\|\|\bar h_{\mathcal N}\|\cos\phi$, we recompute the base-to-GRPO@500 change holding one factor fixed. Closing $\phi$ at base norms gives $15.43\!\to\!8.54$, i.e.\ $84\%$ of the observed drop; moving the norms at base $\phi$ gives $15.43\!\to\!14.46$, i.e.\ $12\%$. The remaining $4\%$ is their interaction.

\paragraph{Which mode moves.}
Displacements from the base centroid, in units of the base gap $\|v_{\mathrm{leak}}\|=15.426$: $\bar h_{\mathcal T}$ moves $5.8\%$ and $10.2\%$ at steps $200$ and $500$, rotating $0.700\degree$ and $1.205\degree$; $\bar h_{\mathcal N}$ moves $53.2\%$ and $66.2\%$, rotating $5.914\degree$ and $7.361\degree$. \think{} is effectively stationary under training while \nothink{} travels two thirds of the way to it, which is what gives the closing gap its direction.

\subsubsection{Construction and fidelity of \cref{fig:vleak-geometry-a}}
\label{app:embedding}

\Cref{fig:vleak-geometry-a} shows the two centroids, $\bar h_{\mathcal T}$ and $\bar h_{\mathcal N}$, and their difference in the same panel.
These vectors live at very different scales: $\bar h_{\mathcal T}$ and $\bar h_{\mathcal N}$ have norm $75$--$77$ while $v_{\mathrm{leak}}$ has norm $7$--$15$, since $\bar h_{\mathcal T}$ and $\bar h_{\mathcal N}$ share a large common mode. 
No orthogonal projection can render both---the plane that preserves $v_{\mathrm{leak}}$ leaves the origin $78.0\degree$ outside it, drawing the centroid norms at $10\%$--$20\%$ of their true length.
We therefore use a approximate two-dimensional embedding rather than a projection. The three $v_{\mathrm{leak}}$ are placed exactly; the three $\bar h_{\mathcal T}$ are drawn as a single point; and the origin is then fitted by least squares to the four centroid norms. 
\Cref{tab:fig-a-fidelity} compares the each drawn vector against its true value in $\mathbb{R}^{4096}$, showing that the plotting errors were kept within a narrow range.

\begin{figure}[h]
    \centering
    \includegraphics[width=0.5\linewidth]{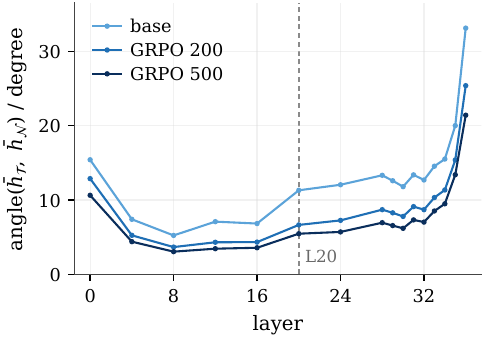}
    \caption{\textbf{The closing gap is a property of training, not of the probe
    layer.} Mode angle $\phi^{(\ell)}$ between $\bar h_{\mathcal T}^{(\ell)}$ and
    $\bar h_{\mathcal N}^{(\ell)}$ across all $16$ probed layers of Qwen3-8B. Dashed line
    marks layer $20$, used throughout the paper.}
    \label{fig:vleak-geometry-b}
\end{figure}

\subsubsection{Depth robustness}

\Cref{fig:vleak-geometry-b} repeats the mode angle $\phi^{(\ell)}$ at $16$ probed layers. The ordering base $>$ GRPO@200 $>$ GRPO@500 holds at every depth, so the closing gap is a property of training and not of the layer at which we probe. The rise beyond layer $32$ is the usual final-layer divergence as the residual stream is prepared for readout.

\subsection{Where methods and models diverge}
\label{app:methods}

The three post-training methods differ in how far their training targets sit from
the model's own distribution: GRPO trains on the model's own samples (on-policy),
OPSD mixes self-generated and external traces (semi-on-policy), and SFT fits
external traces outright (off-policy).

\paragraph{Off-policy training leaks more, and leaks differently.}
Comparing GRPO and SFT checkpoints at matched \nothink{} accuracy, the two
methods separate on both Qwen models: SFT drifts roughly twice as far in paired
effect size and carries about $1.5\times$ the marker density, and on 8B GRPO
reaches slightly higher accuracy ($0.423$ against $0.408$) while drifting less
than half as far ($d=3.29$ against $7.59$). Alignment is high for both
($c\approx0.90$--$0.93$), so the methods move in the same direction and differ
only in how far.

Their trajectories differ as much as their endpoints. SFT has already reached its
final drift at the earliest checkpoint we sample and stays there, whereas GRPO
climbs monotonically across training. The two are consistent with different
origins for the leakage: fitting external traces transfers the \think{} register
wholesale within the first few steps, while on-policy RL accumulates it
gradually, as the samples carrying \think{}-style computation are the ones that
get rewarded. The semi-on-policy middle of the ordering cannot be read---OPSD and
its LoRA variant barely lift accuracy above base, and their small drift is as
consistent with a weak update as with a well-matched one.

\begin{table}[h]
  \centering
  \caption{Style-token density per $1000$ words, counted per word so that the
  three tokenizers are comparable.}
  \label{tab:style-dialect}
  \begin{tabular}{lccc}
    \toprule
    Corpus & $M_{\mathrm{Qwen}}$ & $M_{\mathrm{MiniCPM}}$  & Ratio \\
    \midrule
    OpenThoughts-Math(SFT, OPSD)    & $17.93$ & $10.43$ & $1.72$ \\
    \midrule
    Qwen3-8B \think{}  & $18.81$ & $12.58$ & $1.49$ \\
    Qwen3-4B \think{}  & $19.83$ & $11.35$ & $1.75$ \\
    MiniCPM4.1-8B \think{} & $5.13$ & $18.73$ & $0.27$ \\
    \bottomrule
  \end{tabular}
\end{table}

\paragraph{Style mismatch drives both training failure and reversed drift.}
The corpus that SFT and OPSD train on is written in a measurable style, and that style matches one of the two model families but not the other (\cref{tab:style-dialect}): the traces favour $M_{\mathrm{Qwen}}$ markers over $M_{\mathrm{MiniCPM}}$ ones at a ratio of $1.72$, close to what the two Qwen models produce in their own \think{} rollouts, while MiniCPM4.1-8B produces the opposite balance at $0.27$.

On the Qwen3 models both methods train in the intended direction and both drift toward the $\hat v_{\mathrm{leak}}$, though OPSD does so far less reliably than SFT, which reaches $0.417$ on Qwen3-8B. At matched accuracy SFT drifts about twice as far as GRPO: fitting traces written in these models' own style is an efficient way to
re-invoke \think{}.

On MiniCPM4.1-8B neither method works well. Most SFT and OPSD checkpoints degrade in accuracy and drift away from the $\hat v_{\mathrm{leak}}$. Both effects follow from the mismatch --- the corpus asks the model to reason in a style it does not use. GRPO, trained on the model's own samples, is the only method that succeeds on all three models.

\paragraph{MiniCPM4.1-8B behaves differently from Qwen3 in the first place.}
The style-token balance is one of several ways the two families differ. Response length is another: MiniCPM4.1-8B's \nothink{} answers are $2.5\times$ longer than Qwen3's on the same problems, and its \think{} rollouts $1.8\times$ longer. The architectures differ as well---$32$ layers against Qwen3's $36$, so the middle layer we probe sits at a different relative depth. These base-model differences set the scale of what we measure---how many markers a rollout carries, how far the two mode centroids sit apart, how much a projection can move---so the magnitudes differ across families while the trends do not.

\begin{table}[h]
\centering
\small
\setlength{\tabcolsep}{6pt}
\caption{\textbf{Full steering sweep results.} \emph{Density} is the reasoning-marker
density of \cref{sec:correlational}.
\emph{Trunc.} is the share of rollouts that reach the generation cap and \emph{Loop} the
share whose word-level $12$-gram repetition ratio exceeds $0.3$; the caps and the
repetition threshold are those of \cref{app:setup}.}
\begin{tabular}{llrrrrr}
\toprule
Model & Configuration & Acc. & Density & Trunc. & Loop & Len. \\
 & & & ($\times 10^{-3}$) & (\%) & (\%) & (tok) \\
\midrule
\multirow{8}{*}{Qwen3-8B} & base, \nothink{} & 0.178 & 3.08 & 3.1 & 7.9 & 4,268 \\
 & base, \think{} & 0.557 & 18.30 & 12.4 & 1.1 & 17,782 \\
 & GRPO@500 (best ckpt) & 0.472 & 11.85 & 1.6 & 2.2 & 9,293 \\
\cmidrule(lr){2-7}
 & \multicolumn{6}{l}{\emph{steering, $v_{\mathrm{leak}}$ at $\ell{=}20$}} \\
 & \quad $\alpha=0.5$ & 0.231 & 5.57 & 1.3 & 6.0 & 4,940 \\
 & \quad $\alpha=1$ & 0.335 & 10.26 & 0.8 & 2.9 & 6,850 \\
 & \quad $\alpha=1.25$ & 0.374 & 13.07 & 0.9 & 2.7 & 8,603 \\
 & \quad $\alpha=1.5$ & 0.435 & 15.32 & 1.2 & 2.6 & 10,773 \\
 & \quad $\alpha=2$ & 0.451 & 17.75 & 6.0 & 2.7 & 14,587 \\
\midrule
\multirow{8}{*}{Qwen3-4B} & base, \nothink{} & 0.160 & 3.24 & 2.3 & 7.8 & 4,069 \\
 & base, \think{} & 0.544 & 18.37 & 10.0 & 0.7 & 17,552 \\
 & GRPO@450 (best ckpt) & 0.445 & 10.66 & 0.3 & 1.6 & 8,912 \\
\cmidrule(lr){2-7}
 & \multicolumn{6}{l}{\emph{steering, $v_{\mathrm{leak}}$ at $\ell{=}20$}} \\
 & \quad $\alpha=0.5$ & 0.215 & 5.89 & 1.1 & 7.1 & 4,828 \\
 & \quad $\alpha=1$ & 0.299 & 11.62 & 0.3 & 2.9 & 6,570 \\
 & \quad $\alpha=1.25$ & 0.352 & 13.96 & 2.0 & 2.1 & 8,647 \\
 & \quad $\alpha=1.5$ & 0.396 & 15.60 & 1.0 & 2.1 & 10,234 \\
 & \quad $\alpha=2$ & 0.422 & 16.28 & 4.4 & 0.9 & 14,352 \\
\midrule
\multirow{8}{*}{MiniCPM4.1-8B} & base, \nothink{} & 0.180 & 3.76 & 5.1 & 31.7 & 10,268 \\
 & base, \think{} & 0.584 & 10.55 & 5.8 & 0.4 & 31,430 \\
 & GRPO@500 (best ckpt) & 0.436 & 7.30 & 0.4 & 1.6 & 8,812 \\
\cmidrule(lr){2-7}
 & \multicolumn{6}{l}{\emph{steering, $v_{\mathrm{leak}}$ at $\ell{=}20$}} \\
 & \quad $\alpha=0.5$ & 0.280 & 5.02 & 2.0 & 9.7 & 14,291 \\
 & \quad $\alpha=1$ & 0.398 & 7.21 & 2.8 & 1.2 & 18,315 \\
 & \quad $\alpha=1.25$ & 0.440 & 8.23 & 3.7 & 0.6 & 22,100 \\
 & \quad $\alpha=1.5$ & 0.461 & 9.49 & 4.9 & 0.4 & 25,995 \\
 & \quad $\alpha=2$ & 0.421 & 11.02 & 14.0 & 2.6 & 32,669 \\
\bottomrule
\end{tabular}
\label{tab:steer-sweep-full}
\end{table}

\section{Supplementary Analysis for
  \texorpdfstring{\cref{sec:sufficiency}}{Inducing Leakage by Steering}}
\label{app:steering}

This appendix supports \cref{sec:sufficiency} in three parts: a control that repeats the intervention along the reference axis $\hat v^{(\ell)}_{\text{ref}}$ instead of $v_{\mathrm{leak}}$ (\cref{app:vref-steer}), the same sweep applied at six injection depths (\cref{app:layer-sweep}), and an account of the accuracy drop MiniCPM4.1-8B shows
at the largest dose (\cref{app:degrade}). \Cref{tab:steer-sweep-full} gives the full sweep from which the percentages quoted in the main text are computed.
All tables in this subsection use the column definitions
in \cref{tab:steer-sweep-full}, unless otherwise noted.

\subsection{Control experiment using
  \texorpdfstring{$v_{\mathrm{ref}}$}{reference direction}}
\label{app:vref-steer}
To check that the steering result is specific to $v_{\mathrm{leak}}$, we repeat it on the control
axis $v_{\text{ref}}$ (defined in \cref{eq:vref}), at $\ell{=}20$ with $\alpha{=}1$, on all three
models (\cref{tab:vref-steer}). Note that, on all three models $v_{\text{ref}}$ is the longer vector, so at a common $\alpha$ the control displaces the hidden state further than $v_{\mathrm{leak}}$ does.

On all the three models, neither accuracy nor reasoning-marker density moves off its base value, while the same dose along $v_{\mathrm{leak}}$ lifts both by a wide margin. This confirms leakage runs along $v_{\mathrm{leak}}$. 
Response length does move, and it moves down. This says $v_{\text{ref}}$ is not an empty direction but a meaningful one: it carries the concise clean-solution style of \think{} mode after the exploratory reasoning inside \texttt{<think>}\,$\ldots$\,\texttt{</think>}. Steering along it makes the model write a clean solution without doing the exploration that would normally produce one, so the imitation stays on the surface and accuracy does not rise.

\begin{table}[h]
\centering
\small
\setlength{\tabcolsep}{6pt}
\caption{\textbf{Steering along the reference axis instead of the leakage axis} ($\alpha{=}1$, $\ell{=}20$).}
\begin{tabular}{llrrrrr}
\toprule
Model & Configuration & Acc. & Density & Trunc. & Loop & Len. \\
 & & & ($\times 10^{-3}$) & (\%) & (\%) & (tok) \\
\midrule
\multirow{3}{*}{Qwen3-8B} & base, \nothink{} & 0.178 & 3.08 & 3.1 & 7.9 & 4,268 \\
 & steer along $v_{\mathrm{leak}}$ & 0.335 & 10.26 & 0.8 & 2.9 & 6,850 \\
 & steer along $v_{\text{ref}}$ & 0.159 & 3.10 & 0.1 & 1.9 & 2,057 \\
\midrule
\multirow{3}{*}{Qwen3-4B} & base, \nothink{} & 0.160 & 3.23 & 2.3 & 7.8 & 4,069 \\
 & steer along $v_{\mathrm{leak}}$ & 0.299 & 11.62 & 0.3 & 2.9 & 6,570 \\
 & steer along $v_{\text{ref}}$ & 0.150 & 3.00 & 0.0 & 2.1 & 1,950 \\
\midrule
\multirow{3}{*}{MiniCPM4.1-8B} & base, \nothink{} & 0.180 & 3.76 & 5.1 & 31.7 & 10,268 \\
 & steer along $v_{\mathrm{leak}}$ & 0.398 & 7.21 & 2.8 & 1.2 & 18,315 \\
 & steer along $v_{\text{ref}}$ & 0.148 & 3.57 & 1.4 & 4.0 & 5,463 \\
\bottomrule
\end{tabular}
\label{tab:vref-steer}
\end{table}

\subsection{Steering at other depths}
\label{app:layer-sweep}
The steering results in the main text all inject $v^{(\ell)}_{\text{leak}}$ at $\ell{=}20$ ($v^{(20)}_{\text{leak}}$). As a supplementary
check we ask whether the effect is tied to that depth, or whether the $v^{(\ell)}_{\text{leak}}$ would works at other layers. We repeat the intervention on Qwen3-8B at six depths, $\ell \in \{12, 16, 20, 24, 28, 32\}$, estimating $v^{(\ell)}_{\text{leak}}$
separately at each layer following \cref{eq:vleak} and holding everything else
fixed.

\paragraph{The effect is present at every depth.}
Injecting at any of the six layers raises both \nothink{} accuracy and marker
density above base (\cref{tab:layer-sweep}). The leakage direction is therefore
not an artifact of one probe layer: the same contrast, estimated independently at
each depth, moves the model toward \think{} wherever it is applied.

\paragraph{It is strongest at $\ell{=}20$, on every measure.}
$\ell{=}20$ gains twice as much accuracy as the next best depth, and is the only
one to push marker density past three times the base rate. $\ell{=}20$ is also the most
stable, with the lest degenerate repetitions compared with other layers. Injecting deeper still produces
markers but buys almost no accuracy---the same split between surface signature
and capability that the dose sweep shows, now along depth.  
A coarse scan over the other two models points the same way, with $\ell{=}20$ again the best
depth on both.

\begin{table}[h]
\centering
\small
\setlength{\tabcolsep}{6pt}
\caption{\textbf{Steering the same direction at different depths (Qwen3-8B, $\alpha{=}1$).} $v^{(\ell)}_{\text{leak}}$ is injected at one layer at a time; $^{\dagger}$marks $\ell{=}20$, the layer used everywhere else in the paper. }
\begin{tabular}{lrrrrr}
\toprule
Injection layer & Acc. & Density & Trunc. & Loop & Len. \\
 & & ($\times 10^{-3}$) & (\%) & (\%) & (tok) \\
\midrule
base, \nothink{} & 0.178 & 3.08 & 3.1 & 7.9 & 4,268 \\
\cmidrule(lr){1-6}
$\ell=12$ & 0.254 & 4.25 & 2.1 & 5.6 & 6,527 \\
$\ell=16$ & 0.233 & 4.27 & 1.3 & 7.3 & 5,358 \\
$\ell=20$$^{\dagger}$ & 0.335 & 10.26 & 0.8 & 2.9 & 6,850 \\
$\ell=24$ & 0.224 & 6.68 & 0.5 & 4.2 & 4,260 \\
$\ell=28$ & 0.208 & 6.33 & 1.0 & 8.9 & 4,473 \\
$\ell=32$ & 0.201 & 5.52 & 1.5 & 10.9 & 4,467 \\
\bottomrule
\end{tabular}
\label{tab:layer-sweep}
\end{table}

\subsection{Why MiniCPM4.1-8B loses accuracy at the top dose.}
\label{app:degrade}
MiniCPM4.1-8B is the only model whose accuracy drops at the largest dose, from $0.461$ at
$\alpha{=}1.5$ to $0.421$ at $\alpha{=}2$, while the share of rollouts that hit the
$63488$-token cap rises from $4.9\%$ to $14.0\%$ (\cref{tab:steer-sweep-full}). A truncated
rollout produces no \texttt{\textbackslash boxed} answer and is scored wrong, so we look at
where both fall across questions. 

Truncation is spread out rather than stuck on a few bad
items: at $\alpha{=}2$ it reaches just over half the questions, and on most of those only a
small share of the $16$ rollouts is affected. It does fall mainly on the hard questions,
which average $0.29$ accuracy at $\alpha{=}1.5$ against $0.69$ for the questions with no
truncation. The accuracy drop sits in the same place: on the questions where nothing is
truncated, the two doses score the same, so the whole drop comes from questions where some
rollout ran out of tokens. 

\begin{table}[h]
\centering
\small
\setlength{\tabcolsep}{6pt}
\caption{\textbf{Comparison of steered checkpoint with the base model.} The steered row adds $\alpha{=}0.5, 1, 1.5$ of $v_{\mathrm{leak}}$ to the best GRPO checkpoint of Qwen3-8B.}
\begin{tabular}{lrrrrr}
\toprule
Configuration & Acc. & Density & Trunc. & Loop & Len. \\
 & & ($\times 10^{-3}$) & (\%) & (\%) & (tok) \\
\midrule
base, \nothink{} & 0.178 & 3.08 & 3.1 & 7.9 & 4,268 \\
base, \think{} & 0.557 & 18.30 & 12.4 & 1.1 & 17,782 \\
\cmidrule(lr){1-6}
GRPO@500 & 0.472 & 11.85 & 1.6 & 2.2 & 9,293 \\
\quad $+$ steering, $\alpha{=}0.5$ & 0.545 & 16.42 & 1.3 & 1.8 & 12,179 \\
\quad $+$ steering, $\alpha{=}1$ & 0.553 & 18.31 & 3.8 & 1.0 & 15,033 \\
\quad $+$ steering, $\alpha{=}1.5$ & 0.537 & 18.40 & 7.7 & 0.9 & 17,102 \\
\bottomrule
\end{tabular}
\label{tab:steer-trained-ckpt}
\end{table}

\subsection{Steering a post-trained checkpoint}
\label{app:steer-trained}
The steering experiments in the main text start from the base model. Here we apply the same
operator, at the same layer, to a model that has already been post-trained: Qwen3-8B's best GRPO
checkpoint (\cref{tab:steer-trained-ckpt}).

\paragraph{Post-training does not use up the axis, and the ceiling is \think{} itself.}
Adding $v_{\mathrm{leak}}$ to the checkpoint raises accuracy from $0.472$ to $0.545$ at
$\alpha{=}0.5$ and to $0.553$ at $\alpha{=}1$, where it comes within $0.4$ percents of what the base
model reaches in \think{} mode under the same generation budget, with the marker density arriving at
the same place ($18.31$ against \think{}'s $18.30$). Pushing further does not help: at
$\alpha{=}1.5$ density no longer moves and accuracy falls back. Training and steering therefore
compose rather than substitute, and the base model's \think{} behavior acts as the ceiling---the
direction still has room in it after training, but \nothink{} behavior converges on \think{} rather than
passing it.

\begin{figure}[h]
  \centering
  \includegraphics[width=\linewidth]{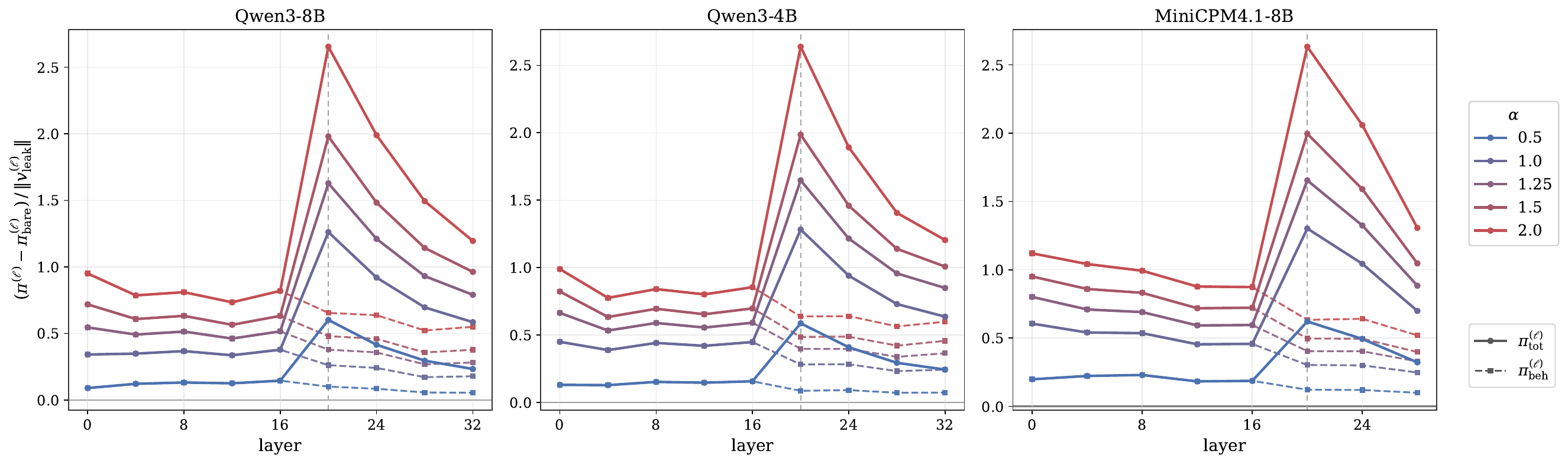}
  \caption{\textbf{Depth profile of steering} ($\alpha = 0.5, 1, 1.25, 1.5, 2$). $\pi^{(\ell)}$ across probe depths, relative to each model's bare
  \nothink{} run and divided by $\|v_{\mathrm{leak}}^{(\ell)}\|$. Colour is the
  steering dose. \emph{Total drift} ($\pi^{(\ell)}_{\mathrm{to t}}$, solid) is measured during steered generation;
  \emph{behavioral drift} ($\pi^{(\ell)}_{\mathrm{beh}}$, dashed) re-forwards the same text without the intervention.
  The dashed vertical line marks the injection layer $\ell{=}20$.}
  \label{fig:depth-profile-steer}
\end{figure}

\subsection{The injected shift versus the shift the text carries}
\label{app:depth-profile-steer}
After the injection, the total drift $\pi^{(\ell)}_{\mathrm{tot}}$ mixes two things: the
steering vector we add, $\alpha\lVert v^{(20)}_{\text{leak}}\rVert$, and the influence of
the steered tokens. To isolate the latter, we re-forward each steered generation without
the intervention, so the resulting profile, $\pi^{(\ell)}_{\mathrm{beh}}$, reflects only
the text.

Both $\pi^{(\ell)}_{\mathrm{tot}}$ and $\pi^{(\ell)}_{\mathrm{beh}}$ grow with dose, but behave differently with depth
(\cref{fig:depth-profile-steer}). The total shift peaks at the injection layer and decays
over the following layers, as later computation absorbs part of the added vector. The
behavioral drift $\pi^{(\ell)}_{\mathrm{beh}}$ is smaller and present at every depth, including layers below the
injection point: re-forwarding adds nothing anywhere, yet passing the steered text through
the model displaces it along $v_{\mathrm{leak}}$ from the first layers on. 

\section{Supplementary Analysis for
  \texorpdfstring{\cref{sec:necessity}}{Removing Leakage by Counter-Steering}}
\label{app:clamp}

This appendix supports \cref{sec:necessity} in four parts. \Cref{app:checkpoint-groups}
sets out the grouping of the fifteen checkpoints on which the analysis rests, and
\cref{app:counter-steer-controls} repeats the intervention along a reference and a random
axis. Four variants of the operator follow: twice the dose, the whole drift rather than
its leakage component, a projection clamp, and the base model in place of a checkpoint
(\cref{app:counter-steer-x2,app:drift-vs-axis,app:clamp-variant,app:counter-steer-base}).
The last two read what the intervention leaves behind---comparing the counter-steered
checkpoint to the base model question by question
(\cref{app:counter-steer-reversion}), and separating the injected shift from the shift the
generated text carries (\cref{app:depth-profile-counter-steer}).
\Cref{tab:counter-steer-full} gives the per-checkpoint results from which the numbers
quoted in the main text are computed.
Throughout this section, all tables use the column definitions in \cref{tab:steer-sweep-full} unless otherwise noted. Shading follows \cref{tab:counter-steer-full} and indicates the checkpoint groups defined in \cref{app:checkpoint-groups}.

\definecolor{grpB}{gray}{0.92}
\definecolor{grpC}{rgb}{0.90,0.94,0.99}
\begin{table}[h]
\centering
\small
\setlength{\tabcolsep}{4.5pt}
\caption{\textbf{Full counter-steering result} ($\ell{=}20$, $\gamma{=}1$). \emph{bare} is the original checkpoint, \emph{c.-s.} the same checkpoint under counter-steering. 
$c=\cos(\delta_s,\hat v_{\mathrm{leak}})$ as defined in \cref{eq:align}.
Rows are grouped as described in \cref{app:checkpoint-groups}: unshaded, the nine \emph{aligned} checkpoints; grey, the four with \emph{no usable gain}; blue, the two \emph{reverse-drift} checkpoints.}
\begin{tabular}{ll r rr rr rr rr}
\toprule
& & & \multicolumn{2}{c}{Accuracy} & \multicolumn{2}{c}{Density} & \multicolumn{2}{c}{Loop (\%)} & \multicolumn{2}{c}{Len. (tok)} \\
\cmidrule(lr){4-5}\cmidrule(lr){6-7}\cmidrule(lr){8-9}\cmidrule(lr){10-11}
Model & Checkpoint & $c$ & bare & c.-s. & bare & c.-s. & bare & c.-s. & bare & c.-s. \\
\midrule
\multirow{5}{*}{Qwen3-8B} & GRPO@500 & +0.95 & 0.472 & 0.349 & 11.85 & 6.51 & 2.2 & 4.5 & 9,293 & 6,991 \\
 & GRPO@200 & +0.95 & 0.447 & 0.315 & 9.47 & 4.55 & 4.3 & 13.9 & 12,583 & 9,936 \\
 & SFT@50 & +0.91 & 0.417 & 0.267 & 13.91 & 9.72 & 2.4 & 14.2 & 13,745 & 12,337 \\
 & OPSD@25 & +0.77 & 0.331 & 0.210 & 6.59 & 2.70 & 4.8 & 28.2 & 13,847 & 9,244 \\
 & \cellcolor{grpB}OPSD@100 & \cellcolor{grpB}+0.46 & \cellcolor{grpB}0.150 & \cellcolor{grpB}0.131 & \cellcolor{grpB}2.33 & \cellcolor{grpB}1.90 & \cellcolor{grpB}23.3 & \cellcolor{grpB}28.0 & \cellcolor{grpB}8,613 & \cellcolor{grpB}6,088 \\
\midrule
\multirow{5}{*}{Qwen3-4B} & GRPO@450 & +0.93 & 0.445 & 0.312 & 10.66 & 5.45 & 1.6 & 6.8 & 8,912 & 7,187 \\
 & GRPO@200 & +0.96 & 0.414 & 0.284 & 9.94 & 5.01 & 0.6 & 12.1 & 10,667 & 8,021 \\
 & SFT@75 & +0.90 & 0.408 & 0.261 & 14.08 & 9.83 & 1.3 & 19.3 & 14,035 & 12,840 \\
 & \cellcolor{grpB}OPSD@50 & \cellcolor{grpB}+0.35 & \cellcolor{grpB}0.191 & \cellcolor{grpB}0.152 & \cellcolor{grpB}2.26 & \cellcolor{grpB}1.83 & \cellcolor{grpB}29.7 & \cellcolor{grpB}34.3 & \cellcolor{grpB}7,299 & \cellcolor{grpB}6,409 \\
 & \cellcolor{grpB}OPSD@25 & \cellcolor{grpB}+0.21 & \cellcolor{grpB}0.119 & \cellcolor{grpB}0.102 & \cellcolor{grpB}1.38 & \cellcolor{grpB}1.14 & \cellcolor{grpB}50.4 & \cellcolor{grpB}51.9 & \cellcolor{grpB}6,068 & \cellcolor{grpB}5,638 \\
\midrule
\multirow{5}{*}{MiniCPM4.1-8B} & GRPO@500 & +0.68 & 0.436 & 0.304 & 7.30 & 5.78 & 1.6 & 6.6 & 8,812 & 6,575 \\
 & GRPO@200 & +0.53 & 0.304 & 0.243 & 6.83 & 4.97 & 9.8 & 25.2 & 8,257 & 7,299 \\
 & \cellcolor{grpC}SFT@100 & \cellcolor{grpC}-0.22 & \cellcolor{grpC}0.162 & \cellcolor{grpC}0.179 & \cellcolor{grpC}5.21 & \cellcolor{grpC}5.50 & \cellcolor{grpC}46.0 & \cellcolor{grpC}38.5 & \cellcolor{grpC}11,091 & \cellcolor{grpC}11,850 \\
 & \cellcolor{grpB}OPSD-LoRA@75 & \cellcolor{grpB}+0.01 & \cellcolor{grpB}0.196 & \cellcolor{grpB}0.189 & \cellcolor{grpB}3.92 & \cellcolor{grpB}4.01 & \cellcolor{grpB}26.8 & \cellcolor{grpB}28.6 & \cellcolor{grpB}10,540 & \cellcolor{grpB}10,662 \\
 & \cellcolor{grpC}OPSD@50 & \cellcolor{grpC}-0.11 & \cellcolor{grpC}0.068 & \cellcolor{grpC}0.085 & \cellcolor{grpC}2.60 & \cellcolor{grpC}2.61 & \cellcolor{grpC}15.1 & \cellcolor{grpC}13.0 & \cellcolor{grpC}12,108 & \cellcolor{grpC}13,360 \\
\bottomrule
\end{tabular}
\label{tab:counter-steer-full}
\end{table}

\subsection{Which checkpoints the analysis uses}
\label{app:checkpoint-groups}
Counter-steering removes a checkpoint's drift along $v_{\mathrm{leak}}$, so what it can tell us
depends on what training put there. The fifteen checkpoints fall into three groups on two
measurements that are fixed before any intervention is run: how far the checkpoint's accuracy rose
above its \nothink{} base, and how the training drift is oriented with respect to $v_{\mathrm{leak}}$
(\cref{tab:counter-steer-full}).

\paragraph{Aligned (nine checkpoints).}
Training moved these models a long way onto the leakage axis and made them better \nothink{}
solvers ($\cos \geq 0.53$, gain $\geq +0.12$). Both a gain and a mechanism that could carry it are
present, so these are the checkpoints the analysis in \cref{sec:necessity} mainly uses.

\paragraph{No usable gain (four checkpoints).}
Training left these models no better than their base, whether or not it moved them along the axis
($|\text{gain}| \leq 0.04$, $0.01 \leq \cos \leq 0.46$). With no gain there is nothing whose
drift-induced gain can be tested, and no denominator to normalize a loss against, so they are excluded.

\paragraph{Reverse drift (two checkpoints).}
Training moved these models \emph{reversely} on the leakage axis, and they did not improve
($\cos \leq -0.11$, gain $\leq -0.02$). These are used as a complementary group of the aligned group.

\subsection{Control experiment: reference and random axes}
\label{app:counter-steer-controls}

To confirm that the counter-steering result of \cref{sec:necessity} is specific to
$v_{\mathrm{leak}}$, we repeat it on each model's best checkpoint along two control axes:
the reference axis $\hat v_{\text{ref}}$ of \cref{app:vref}, and a direction drawn at random per question.  
The dosing rule is unchanged---$\gamma{=}1$ subtracts the checkpoint's drift component
along whichever axis is used---so the control dose is not chosen by us but read off the
drift itself, and a null result can arise two ways: because training put nothing on that
axis, or because what it put there does not matter.

\begin{table}[t]
\centering
\small
\setlength{\tabcolsep}{6pt}
\caption{\textbf{Counter-steering along the leakage axis and two control axes} ($\ell{=}20$, $\gamma{=}1$, best checkpoint per model). }
\begin{tabular}{llrrrrr}
\toprule
Model & Configuration & Acc. & Density & Trunc. & Loop & Len. \\
 & & & ($\times 10^{-3}$) & (\%) & (\%) & (tok) \\
\midrule
\multirow{4}{*}{Qwen3-8B} & GRPO@500 & 0.472 & 11.85 & 1.6 & 2.2 & 9,293 \\
\cmidrule(lr){2-7}
 & \multicolumn{6}{l}{\emph{counter-steering}} \\
 & \quad $v_{\mathrm{leak}}$ & 0.349 & 6.51 & 1.2 & 4.5 & 6,991 \\
 & \quad $v_{\text{ref}}$ & 0.459 & 12.06 & 0.5 & 2.1 & 9,090 \\
 & \quad random & 0.455 & 12.05 & 0.9 & 1.9 & 9,325 \\
\midrule
\multirow{4}{*}{Qwen3-4B} & GRPO@450 & 0.445 & 10.66 & 0.3 & 1.6 & 8,912 \\
\cmidrule(lr){2-7}
 & \multicolumn{6}{l}{\emph{counter-steering}} \\
 & \quad $v_{\mathrm{leak}}$ & 0.312 & 5.45 & 1.0 & 6.8 & 7,187 \\
 & \quad $v_{\text{ref}}$ & 0.435 & 10.61 & 0.4 & 1.4 & 9,097 \\
 & \quad random & 0.430 & 10.64 & 0.5 & 1.9 & 9,145 \\
\midrule
\multirow{4}{*}{MiniCPM4.1-8B} & GRPO@500 & 0.436 & 7.30 & 0.4 & 1.6 & 8,812 \\
\cmidrule(lr){2-7}
 & \multicolumn{6}{l}{\emph{counter-steering}} \\
 & \quad $v_{\mathrm{leak}}$ & 0.304 & 5.78 & 0.4 & 6.6 & 6,575 \\
 & \quad $v_{\text{ref}}$ & 0.451 & 7.36 & 0.8 & 1.5 & 9,119 \\
 & \quad random & 0.449 & 7.24 & 0.3 & 1.4 & 8,807 \\
\bottomrule
\end{tabular}
\label{tab:counter-steer-controls}
\end{table}

\paragraph{On the Qwen models, training put nothing there.}
Along $\hat v_{\text{ref}}$ the drift components are $0.07$ and $0.001$, against drift
lengths of $10.52$ and $5.57$ and below the $0.16$ and $0.11$ a random direction picks
up by chance (\cref{tab:counter-steer-controls}). The prescribed dose is therefore
negligible, and both controls leave accuracy and marker density where they were.

\paragraph{On MiniCPM4.1-8B the dose is real, and still costs nothing.}
Here the $\hat v_{\text{ref}}$ component is about half the $v_{\mathrm{leak}}$ one, so the
control is a substantive intervention rather than a no-op. It leaves accuracy and marker
density unchanged, while removing the $v_{\mathrm{leak}}$ component of the same drift
costs roughly a third of the accuracy. This is the stronger form of the control: what a
checkpoint gains in training sits on $v_{\mathrm{leak}}$, and displacing it by a
comparable amount along another meaningful axis does not touch the gain.


\subsection{Doubling the dose}
\label{app:counter-steer-x2}
Counter-steering at $\gamma{=}1$ nominally returns a checkpoint to the base engagement level. At
$\gamma{=}2$ it travels the same distance again, landing as far below $\pi_0$ as the
checkpoint sat above it.

\begin{table}[t]
\centering
\small
\setlength{\tabcolsep}{5pt}
\caption{\textbf{Comparison of counter-steering at $\times1$ and $\times2$ dose} ($\gamma{=}1, 2$).}
\begin{tabular}{ll rr rr rr rr}
\toprule
& & \multicolumn{2}{c}{Accuracy} & \multicolumn{2}{c}{Density} & \multicolumn{2}{c}{Loop (\%)} & \multicolumn{2}{c}{Len. (tok)} \\
\cmidrule(lr){3-4}\cmidrule(lr){5-6}\cmidrule(lr){7-8}\cmidrule(lr){9-10}
Model & Checkpoint & $\times 1$ & $\times 2$ & $\times 1$ & $\times 2$ & $\times 1$ & $\times 2$ & $\times 1$ & $\times 2$ \\
\midrule
\multirow{5}{*}{Qwen3-8B} & GRPO@500 & 0.349 & 0.212 & 6.51 & 3.43 & 4.5 & 13.3 & 6,991 & 6,078 \\
 & GRPO@200 & 0.315 & 0.191 & 4.55 & 2.40 & 13.9 & 22.6 & 9,936 & 7,696 \\
 & SFT@50 & 0.267 & 0.131 & 9.72 & 4.79 & 14.2 & 30.1 & 12,337 & 9,587 \\
 & OPSD@25 & 0.210 & 0.087 & 2.70 & 1.21 & 28.2 & 59.6 & 9,244 & 6,154 \\
 & \cellcolor{grpB}OPSD@100 & \cellcolor{grpB}0.131 & \cellcolor{grpB}0.076 & \cellcolor{grpB}1.90 & \cellcolor{grpB}1.53 & \cellcolor{grpB}28.0 & \cellcolor{grpB}41.4 & \cellcolor{grpB}6,088 & \cellcolor{grpB}5,353 \\
\midrule
\multirow{5}{*}{Qwen3-4B} & GRPO@450 & 0.312 & 0.212 & 5.45 & 2.82 & 6.8 & 12.0 & 7,187 & 5,799 \\
 & GRPO@200 & 0.284 & 0.191 & 5.01 & 2.74 & 12.1 & 15.9 & 8,021 & 6,387 \\
 & SFT@75 & 0.261 & 0.118 & 9.83 & 4.66 & 19.3 & 36.5 & 12,840 & 9,776 \\
 & \cellcolor{grpB}OPSD@50 & \cellcolor{grpB}0.152 & \cellcolor{grpB}0.119 & \cellcolor{grpB}1.83 & \cellcolor{grpB}1.31 & \cellcolor{grpB}34.3 & \cellcolor{grpB}39.4 & \cellcolor{grpB}6,409 & \cellcolor{grpB}5,525 \\
 & \cellcolor{grpB}OPSD@25 & \cellcolor{grpB}0.102 & \cellcolor{grpB}0.086 & \cellcolor{grpB}1.14 & \cellcolor{grpB}1.14 & \cellcolor{grpB}51.9 & \cellcolor{grpB}56.0 & \cellcolor{grpB}5,638 & \cellcolor{grpB}5,174 \\
\midrule
\multirow{5}{*}{MiniCPM4.1-8B} & GRPO@500 & 0.304 & 0.228 & 5.78 & 4.69 & 6.6 & 29.3 & 6,575 & 6,349 \\
 & GRPO@200 & 0.243 & 0.187 & 4.97 & 4.27 & 25.2 & 44.5 & 7,299 & 7,248 \\
 & \cellcolor{grpC}SFT@100 & \cellcolor{grpC}0.179 & \cellcolor{grpC}0.201 & \cellcolor{grpC}5.50 & \cellcolor{grpC}6.34 & \cellcolor{grpC}38.5 & \cellcolor{grpC}33.4 & \cellcolor{grpC}11,850 & \cellcolor{grpC}12,788 \\
 & \cellcolor{grpB}OPSD-LoRA@75 & \cellcolor{grpB}0.189 & \cellcolor{grpB}0.191 & \cellcolor{grpB}4.01 & \cellcolor{grpB}4.04 & \cellcolor{grpB}28.6 & \cellcolor{grpB}27.5 & \cellcolor{grpB}10,662 & \cellcolor{grpB}10,451 \\
 & \cellcolor{grpC}OPSD@50 & \cellcolor{grpC}0.085 & \cellcolor{grpC}0.098 & \cellcolor{grpC}2.61 & \cellcolor{grpC}2.64 & \cellcolor{grpC}13.0 & \cellcolor{grpC}8.9 & \cellcolor{grpC}13,360 & \cellcolor{grpC}13,355 \\
\bottomrule
\end{tabular}
\label{tab:counter-steer-nf2}
\end{table}

\paragraph{Each group extends its own trend.}
Doubling the dose continues what $\gamma{=}1$ started
(\cref{tab:counter-steer-nf2}). In the aligned group accuracy and marker density fall
again, ending close to the checkpoints' own \nothink{} base on both measures. The
reverse-drift group moves the other way at both doses: these checkpoints drifted backward
along $v_{\mathrm{leak}}$, so subtracting their drift adds the direction rather than
removing it, and accuracy rises with dose.

This gives a two-sided control on the intervention. Most checkpoints drifted toward
\think{} along $v_{\mathrm{leak}}$, and counter-steering subtracts that drift, so it
removes the direction; a few drifted backward, and subtracting their negative drift adds
the direction instead. The operator is the same in both cases. If the accuracy change
came from disrupting the model, both groups would get worse. Instead the aligned group
loses accuracy and the reverse-drift group gains it, each following the sign of its own
drift.

\paragraph{Why we do not use $\gamma{=}2$ quantitatively.}
Repetition climbs steeply in the aligned group and ends above the base model the
checkpoints were pushed back to, so part of the accuracy drop here is degenerate
generation rather than removed leakage, and the two cannot be separated. (In the
reverse-drift group repetition falls with dose, so the rise is not a mechanical
consequence of adding more dose.) Second, $\gamma{=}2$ is try to reduce $\pi$ below base \nothink{} level, where the base model no longer serves as a reference point: at $\gamma{=}1$
we can ask whether accuracy returns to base, but past it the state is one no version of
the model ever generated from. We therefore read this dose as evidence that the effect
is graded and direction-dependent, and base no quantitative claim on it.

\begin{table}[h]
\centering
\small
\setlength{\tabcolsep}{6pt}
\caption{\textbf{Removing the $\hat v_{\mathrm{leak}}$ component of the drift against removing the whole drift} (Qwen3-8B, $\ell{=}20$). The $\hat v_{\mathrm{leak}}$ arm is the counter-steering used throughout the paper, which removes $\Delta\pi\, v_{\mathrm{leak}}$ with $\gamma = 1$; the $\delta_s$ arm removes the drift in full. }
\begin{tabular}{lrrrrr}
\toprule
Configuration & Acc. & Density & Trunc. & Loop & Len. \\
 & & ($\times 10^{-3}$) & (\%) & (\%) & (tok) \\
\midrule
GRPO@500 & 0.472 & 11.85 & 1.6 & 2.2 & 9,293 \\
\quad $-$ $\hat v_{\mathrm{leak}}$ & 0.349 & 6.51 & 1.2 & 4.5 & 6,991 \\
\quad $-$ $\delta_s$ & 0.325 & 6.18 & 1.6 & 5.6 & 6,571 \\
\midrule
GRPO@200 & 0.447 & 9.47 & 6.2 & 4.3 & 12,583 \\
\quad $-$ $\hat v_{\mathrm{leak}}$ & 0.315 & 4.55 & 5.2 & 13.9 & 9,936 \\
\quad $-$ $\delta_s$ & 0.308 & 4.75 & 3.9 & 9.4 & 8,945 \\
\bottomrule
\end{tabular}
\label{tab:drift-vs-axis}
\end{table}

\subsection{How much of the drift the leakage axis accounts for}
\label{app:drift-vs-axis}
Counter-steering using $\Delta\pi\, v_{\mathrm{leak}}$ removes only the component of
the drift that lies along $\hat v_{\mathrm{leak}}$, and \cref{fig:vleak-geometry-a} shows there is
another component orthogonal to it. The drift
$\delta_s = \bar h_{\mathcal{N}}(\theta_s) - \bar h_{\mathcal{N}}(\theta_0)$
decomposes as $\delta_s = \Delta\pi\, v_{\mathrm{leak}} + v_{\perp}$. To check
whether $v_{\perp}$ matters, we remove $\delta_s$ in full instead.

\paragraph{The axis accounts for almost all of it.}
As shown in \cref{tab:drift-vs-axis}, removing $\delta_s$ in full reduces only slightly more accuracy
than removing its leakage-axis component alone, so the axis captures $84\%$ and $95\%$ of what the
full intervention achieves. The axis estimated on the base model therefore remains the right one
after post-training.

\begin{table}[h]
\centering
\small
\setlength{\tabcolsep}{7pt}
\caption{\textbf{Counter-steering versus the projection clamp}, averaged over the $15$
checkpoints of \cref{sec:necessity}. Both set $\gamma{=}1$.}
\begin{tabular}{lrrrrr}
\toprule
Configuration & Acc. & Density & Trunc. & Loop & Len. \\
 & & ($\times 10^{-3}$) & (\%) & (\%) & (tok) \\
\midrule
base, \nothink{} & 0.173 & 3.36 & 3.5 & 15.8 & 6,202 \\
post-trained checkpoint & 0.304 & 7.22 & 4.9 & 14.7 & 10,391 \\
\cmidrule(lr){1-6}
\quad + counter-steering & 0.226 & 4.77 & 3.7 & 21.7 & 8,963 \\
\quad + projection clamp & 0.240 & 5.84 & 3.6 & 24.5 & 9,220 \\
\bottomrule
\end{tabular}
\label{tab:counter-steer-summary}
\end{table}

\subsection{Why counter-steering, and not a projection clamp}
\label{app:clamp-variant}

Counter-steering shifts the projection by a constant, which sets the mean
engagement but leaves each position free to depart from it. The operation that
realises $\mathrm{do}(\pi{=}\pi_0)$ exactly is to pin the coordinate at every
position instead,
\begin{equation}
h^{(\ell)}_t \;\leftarrow\; h^{(\ell)}_t
- \Big(\big\langle h^{(\ell)}_t,\hat v^{(\ell)}_{\text{leak}}\big\rangle
- \pi_0^{(\ell)}\Big)\hat v^{(\ell)}_{\text{leak}} ,
\label{eq:pin}
\end{equation}
which forces the projection to the base level exactly, and removes the token-to-token variation in engagement
along with it.

\paragraph{Counter-steering removes more and breaks less.}
At the same dose, counter-steering gives both lower marker density and lower accuracy than the clamp, and produces fewer degenerate repetitions (\cref{tab:counter-steer-summary}). We attribute the clamp's shortfall to the variance it flattens: holding every position at one projection costs coherence rather than leaked computation. Neither operator is clean---counter-steering also raises the repetition rate above the original checkpoint, so part of the accuracy it removes is disruption rather than leakage. 
The difference suggests the measurement is still limited by the operator: a finer intervention, one that reaches the same engagement level while staying within states the model could have produced on its own, should remove more of the leaked behavior at less collateral cost, and so read the leakage level more accurately.

\subsection{Counter-steering the base model}
\label{app:counter-steer-base}
Applying counter-steering to the base models reverses the pattern seen under base
steering: accuracy, marker density and response length all fall, and the loop rate
rises (\cref{tab:counter-steer-base}).

\paragraph{The response below base is much weaker than above it.}
Per unit of imposed $\Delta\pi\, v_{\mathrm{leak}}$, base counter-steering loses $0.064$, $0.060$
and $0.109$ on Qwen3-8B, Qwen3-4B and MiniCPM4.1-8B---roughly half the $0.147$,
$0.133$ and $0.215$ that base steering gains per unit in the opposite direction,
and half the $0.166$, $0.178$ and $0.212$ that counter-steering costs a
checkpoint. Moving the base model below its own $\pi_0$ buys far less change than
moving it above.

\begin{table}[h]
\centering
\small
\setlength{\tabcolsep}{6pt}
\caption{\textbf{Counter-steering the base model} ($\ell{=}20$). The magnitude is set as $\gamma\|v_{\mathrm{leak}}\|$ since there is no drift for a base model.
}
\begin{tabular}{llrrrrr}
\toprule
Model & Configuration & Acc. & Density & Trunc. & Loop & Len. \\
 & & & ($\times 10^{-3}$) & (\%) & (\%) & (tok) \\
\midrule
\multirow{3}{*}{Qwen3-8B} & base, \nothink{} & 0.178 & 3.08 & 3.1 & 7.9 & 4,268 \\
 & \quad $\gamma=0.5$ & 0.151 & 1.99 & 1.3 & 8.6 & 3,322 \\
 & \quad $\gamma=1$ & 0.111 & 1.33 & 1.9 & 9.5 & 3,233 \\
\midrule
\multirow{3}{*}{Qwen3-4B} & base, \nothink{} & 0.160 & 3.24 & 2.3 & 7.8 & 4,069 \\
 & \quad $\gamma=0.5$ & 0.129 & 2.02 & 1.0 & 10.4 & 3,270 \\
 & \quad $\gamma=1$ & 0.100 & 1.48 & 0.8 & 10.1 & 2,877 \\
\midrule
\multirow{3}{*}{MiniCPM4.1-8B} & base, \nothink{} & 0.180 & 3.76 & 5.1 & 31.7 & 10,268 \\
 & \quad $\gamma=0.5$ & 0.119 & 3.14 & 1.6 & 54.9 & 8,209 \\
 & \quad $\gamma=1$ & 0.074 & 2.71 & 1.7 & 70.9 & 7,620 \\
\bottomrule
\end{tabular}
\label{tab:counter-steer-base}
\end{table}

\subsection{What counter-steering moves the model back to}
\label{app:counter-steer-reversion}
Counter-steering costs accuracy and adds repetition, and a larger dose does more of
both. Two readings fit that: the intervention is damaging the model in a new way,
or it is undoing what training did and returning it toward the base model. Since
every question is answered $16$ times, we can tell these apart question by
question: if the intervention is damaging, its per-question pattern should follow
neither the checkpoint nor the base model; if it is undoing training, the pattern
should shift from the checkpoint toward the base model as the dose grows.

\paragraph{The repetition pattern resembles the base model's.}
Which questions repeat under counter-steering is predicted by which questions
repeat in the base model ($\rho=+0.52$ at $\gamma{=}1$, $+0.64$ at $\gamma{=}2$),
and the match is closer at the larger dose. The two sets also overlap: $84\%$ of
the questions that repeat under counter-steering also repeat in the base model.

\paragraph{Accuracy moves the same way, more slowly.}
Base and checkpoint question-level accuracy are themselves correlated across questions
($\rho=+0.76$), since both follow how hard each question is, so we compare using
partial correlations that hold one of them fixed. At $\gamma{=}1$ the
counter-steered accuracy stays closer the checkpoint more than the base model
($+0.78$ against $+0.43$); at $\gamma{=}2$ the two are level ($+0.50$ against
$+0.58$). Accuracy moves toward the base model as the dose grows, but not as far
as repetition does.

\paragraph{The repetition does not drive the accuracy loss.}
Keeping only rollouts that terminate normally, accuracy still falls from $0.459$
to $0.342$ at $\gamma{=}1$ and to $0.269$ at $\gamma{=}2$, which is $91\%$ and
$82\%$ of the full loss. What counter-steering removes is therefore not just the
ability to finish a response. The exception is MiniCPM4.1-8B at $\gamma{=}2$,
where repetition reaches $29\%$ of rollouts and accounts for two fifths of the
loss; there the accuracy result cannot be read on its own.

\begin{figure}[h]
  \centering
  \includegraphics[width=\linewidth]{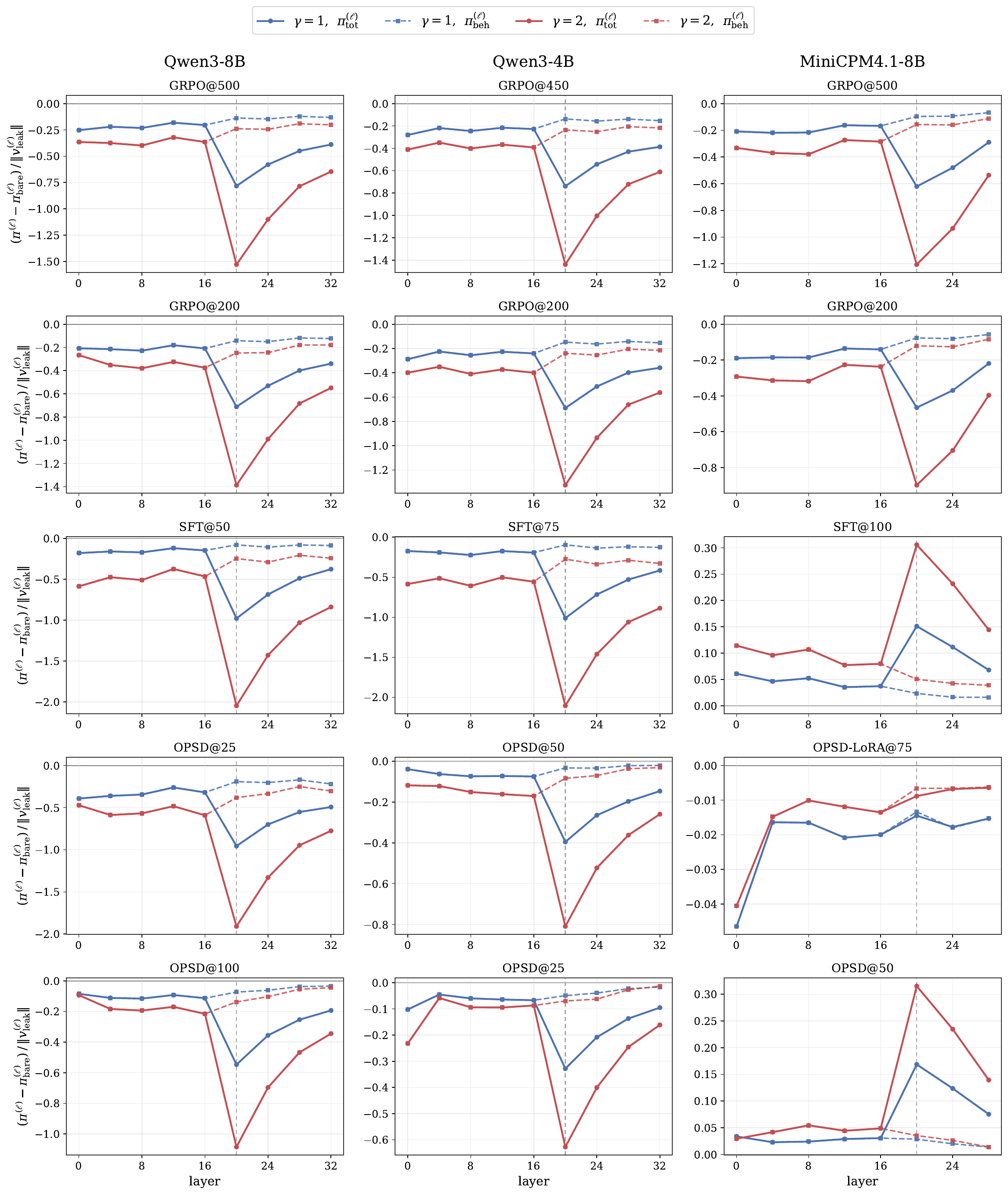}
  \caption{\textbf{Depth profile of counter-steering, checkpoint by checkpoint} ($\gamma = 1, 2$). Columns are
models, rows their five checkpoints. In each panel, $\pi^\ell$ across probe depths, relative to that checkpoint's original run and divided by $\|v_{\mathrm{leak}}^{(\ell)}\|$. Blue is $\gamma{=}1$, red $\gamma{=}2$;
\emph{Total drift} ($\pi^{(\ell)}_{\mathrm{tot}}$, solid) is measured during counter-steered generation;
  \emph{behavioral drift} ($\pi^{(\ell)}_{\mathrm{beh}}$, dashed) re-forwards the same text without the intervention.
  The dashed vertical line marks the injection layer $\ell{=}20$.}
  \label{fig:depth-profile-counter-steer}
\end{figure}

\subsection{The injected shift versus the shift the text carries}
\label{app:depth-profile-counter-steer}
\Cref{fig:depth-profile-counter-steer} repeats the decomposition of
\cref{app:depth-profile-steer} on the counter-steering side, per checkpoint. As defined in~\cref{app:depth-profile-steer}, $\pi^{(\ell)}_{\mathrm{tot}}$ denotes the total drift and $\pi^{(\ell)}_{\mathrm{beh}}$ is the behavioral drift. The picture
mirrors it: the two curves (solid and dashed) coincide below the injection layer, the gap that opens there
closes with depth, and $\pi^{(\ell)}_{\mathrm{beh}}$ stays away from zero throughout, so
part of the engagement change is again carried by the tokens.

The panels also split by the sign of $\Delta\pi$. Where a checkpoint drifted backward
during training, the operator adds $v_{\mathrm{leak}}$ rather than subtracting it, and both
curves move up instead of down. 
These are the same checkpoints whose accuracy rises rather than falls under the operator
(\cref{app:counter-steer-x2}).

\section{Supplementary Analysis for
  \texorpdfstring{\cref{sec:gain_audit}}{Attributing the Gain to Thinking Leakage}}
\label{app:gain_audit}

This appendix supplements \cref{sec:gain_audit} in five parts.
\Cref{app:detail_decomposition} reports the per-checkpoint gain
decompositions and interprets the leakage ratio across checkpoint
groups.
\Cref{app:question-level} examines whether the aggregate effects
align at the question level. \Cref{app:methods-decomposition}
extends the training method comparison using the decomposition.
\Cref{app:response-asymmetry} proposes an asymmetric response profile that would account
for the steering and counter-steering slopes, and \cref{app:three-way} relates the
bidirectional audit to a three-way mediation decomposition as a theoretical extension.

\begin{table}[h]
\centering
\small
\setlength{\tabcolsep}{4.5pt}
\caption{\textbf{Per-checkpoint gain decompositions} ($\ell=20$).
Notation follows \cref{sec:gain_audit}.
}
\begin{tabular}{ll rr rr rr}
\toprule
& & \multicolumn{2}{c}{via the base sweep} & \multicolumn{2}{c}{via counter-steering} & & \\
\cmidrule(lr){3-4}\cmidrule(lr){5-6}
Model & Checkpoint & $L_{\mathrm{base}}$ & $R_1$ & $L_{\mathrm{trained}}$ & $R_0$ & $\mathcal{A}$ & $\lambda$ \\[2pt]
\midrule
\multirow{5}{*}{Qwen3-8B} & GRPO@500 & +0.095 & +0.199 & +0.123 & +0.171 & +0.028 & 0.418 \\
 & GRPO@200 & +0.084 & +0.185 & +0.132 & +0.136 & +0.049 & 0.492 \\
 & SFT@50 & +0.132 & +0.107 & +0.150 & +0.089 & +0.018 & 0.627 \\
 & OPSD@25 & +0.112 & +0.041 & +0.121 & +0.032 & +0.009 & 0.789 \\
 & \cellcolor{grpB}OPSD@100 & \cellcolor{grpB}+0.070 & \cellcolor{grpB}-0.098 & \cellcolor{grpB}+0.019 & \cellcolor{grpB}-0.047 & \cellcolor{grpB}-0.051 & \cellcolor{grpB}-0.667 \\
\midrule
\multirow{5}{*}{Qwen3-4B} & GRPO@450 & +0.080 & +0.205 & +0.133 & +0.153 & +0.053 & 0.465 \\
 & GRPO@200 & +0.072 & +0.182 & +0.130 & +0.124 & +0.057 & 0.510 \\
 & SFT@75 & +0.122 & +0.126 & +0.147 & +0.102 & +0.025 & 0.591 \\
 & \cellcolor{grpB}OPSD@50 & \cellcolor{grpB}+0.048 & \cellcolor{grpB}-0.018 & \cellcolor{grpB}+0.039 & \cellcolor{grpB}-0.008 & \cellcolor{grpB}-0.009 & \cellcolor{grpB}1.271 \\
 & \cellcolor{grpB}OPSD@25 & \cellcolor{grpB}+0.037 & \cellcolor{grpB}-0.078 & \cellcolor{grpB}+0.017 & \cellcolor{grpB}-0.058 & \cellcolor{grpB}-0.021 & \cellcolor{grpB}-0.405 \\
\midrule
\multirow{5}{*}{MiniCPM4.1-8B} & GRPO@500 & +0.113 & +0.144 & +0.133 & +0.124 & +0.020 & 0.517 \\
 & GRPO@200 & +0.084 & +0.041 & +0.061 & +0.063 & -0.022 & 0.494 \\
 & \cellcolor{grpC}SFT@100 & \cellcolor{grpC}-0.027 & \cellcolor{grpC}+0.010 & \cellcolor{grpC}-0.017 & \cellcolor{grpC}-0.001 & \cellcolor{grpC}+0.011 & \cellcolor{grpC}0.941 \\
 & \cellcolor{grpB}OPSD-LoRA@75 & \cellcolor{grpB}+0.000 & \cellcolor{grpB}+0.016 & \cellcolor{grpB}+0.007 & \cellcolor{grpB}+0.009 & \cellcolor{grpB}+0.007 & \cellcolor{grpB}0.419 \\
 & \cellcolor{grpC}OPSD@50 & \cellcolor{grpC}-0.030 & \cellcolor{grpC}-0.082 & \cellcolor{grpC}-0.018 & \cellcolor{grpC}-0.094 & \cellcolor{grpC}+0.012 & \cellcolor{grpC}0.158 \\
\bottomrule
\end{tabular}
\label{tab:gain-decomposition}
\end{table}

\subsection{Details of auditing on drift-induced gain}
\label{app:detail_decomposition}

\Cref{tab:gain-decomposition} gives the per-checkpoint decomposition along both
paths. We read it by the three groups of \cref{sec:necessity}, which are fixed
before any intervention is run.

\paragraph{Aligned checkpoints.}
Both terms are positive throughout: the training improved \nothink{} accuracy
($\Delta>0$), part of that improvement is induced by drift on $\hat{v}_{\mathrm{leak}}$
($L_{\mathrm{trained}}>0$), and part survives it ($R_0>0$). 
Thus $\lambda$ here takes
its plainest reading, the fraction of drift-induced gain, and runs from $0.42$ to $0.79$. The ordering
follows the training method: the GRPO checkpoints sit near $0.5$, SFT near $0.6$,
and OPSD@25 on Qwen3-8B highest at $0.79$.

\paragraph{No usable gain.}
The three Qwen OPSD checkpoints in this group share a pattern:
$L_{\mathrm{trained}}$ is positive while $R_0$ is negative. Counter-steering still
removes accuracy, so the drift did carry a genuine effect, but the improvement outside the drift is negative. Since $\Delta=L_{\mathrm{trained}}+R_0$ is then small or
negative, $\lambda$ falls outside $[0,1]$ and is no longer meaningful. 
The remaining checkpoint, OPSD-LoRA@75 on MiniCPM4.1-8B, has both
terms positive but both within noise ($L_{\mathrm{trained}}=+0.007$,
$R_0=+0.009$): that run barely moved the model in any direction.

\paragraph{Reverse drift.}
Both terms are negative. The negative $L_{\mathrm{trained}}$ says the reverse drift decrease accuracy, and counter-steering recovers the decreasing. The negative $R_0$ says the rest of the training hurt as
well. Because both change sign together with $\Delta$, $\lambda$ stays in $[0,1]$
and keeps its meaning: on MiniCPM4.1-8B SFT@100 it reads $0.94$, so almost all of that checkpoint's degradation is explained by the reverse drift.

\subsection{Question-level alignment}
\label{app:question-level}

In this subsection, we examine whether \think{}, post-training, and the two interventions affect similar questions, beyond their aggregate accuracy effects.

For each of the $120$ questions, let $a_{w,c}(q)$ denote mean
correctness over $16$ rollouts, where $w=0,1$ denotes the base
model or trained checkpoint and $c$ specifies the mode or intervention.
We define
\begin{equation}
\begin{aligned}
g_{\mathrm{T}}(q)
&= a_{0,\think}(q)-a_{0,\nothink}(q),\\
g_{\mathrm{S}}(q)
&= a_{0,\mathrm{steer}}(q)-a_{0,\nothink}(q),\\
\Delta(q)
&= a_{1,\nothink}(q)-a_{0,\nothink}(q),\\
L_{\mathrm{trained}}(q)
&= a_{1,\nothink}(q)-a_{1,\mathrm{counter}}(q).
\end{aligned}
\end{equation}
Here, steering uses the fixed dose $\alpha=1$, while counter-steering
uses the checkpoint-specific full dose $\gamma=1$.
\Cref{tab:question-rho} reports Spearman correlations across questions.

\paragraph{Question-level agreement across interventions.}
Across the nine aligned checkpoints, post-training gains $\Delta$ correlate positively with thinking gains $g_{\mathrm{T}}$ ($\rho=0.54$--$0.89$), steering gains $g_{\mathrm{S}}$ ($\rho=0.57$--$0.79$), and counter-steering losses $L_{\mathrm{trained}}$ ($\rho=0.59$--$0.82$).
Questions that benefit more from post-training therefore tend
to benefit more from \think{} and steering, and to lose more
under counter-steering.
Non-aligned checkpoints show weaker correlations involving
post-training gains, consistent with their limited training
effectiveness.
Together with the positive alignment between thinking and
steering gains ($\rho(g_{\mathrm{T}},g_{\mathrm{S}})=0.61$--$0.73$),
these results extend the aggregate findings in
\cref{sec:correlational,sec:sufficiency,sec:necessity,sec:gain_audit}
to individual questions.

\begin{table}[h]
\centering
\small
\setlength{\tabcolsep}{6pt}
\caption{\textbf{Question-level alignment of gains and intervention
effects.} Spearman correlations across $120$ questions.
The base-only correlation $\rho(g_{\mathrm{T}},g_{\mathrm{S}})$
is identical within each model.
Shading follows \cref{tab:counter-steer-full}.}
\begin{tabular}{ll rrrr}
\toprule
Model & Checkpoint & 
$\rho(g_{\mathrm{S}},\Delta)$&
$\rho(\Delta,L_{\mathrm{trained}})$&
$\rho(g_{\mathrm{T}},\Delta)$&
$\rho(g_{\mathrm{T}},g_{\mathrm{S}})$\\
\midrule
\multirow{5}{*}{Qwen3-8B} & GRPO@500 & +0.772 & +0.766 & +0.836 & \multirow{5}{*}{+0.677} \\
 & GRPO@200 & +0.792 & +0.806 & +0.852 &  \\
 & SFT@50 & +0.754 & +0.752 & +0.882 &  \\
 & OPSD@25 & +0.572 & +0.643 & +0.677 &  \\
 & \cellcolor{grpB}OPSD@100 & \cellcolor{grpB}+0.341 & \cellcolor{grpB}+0.510 & \cellcolor{grpB}+0.349 &  \\
\midrule
\multirow{5}{*}{Qwen3-4B} & GRPO@450 & +0.741 & +0.820 & +0.874 & \multirow{5}{*}{+0.613} \\
 & GRPO@200 & +0.730 & +0.798 & +0.837 &  \\
 & SFT@75 & +0.677 & +0.818 & +0.885 &  \\
 & \cellcolor{grpB}OPSD@50 & \cellcolor{grpB}+0.289 & \cellcolor{grpB}+0.503 & \cellcolor{grpB}+0.274 &  \\
 & \cellcolor{grpB}OPSD@25 & \cellcolor{grpB}-0.253 & \cellcolor{grpB}+0.237 & \cellcolor{grpB}-0.085 &  \\
\midrule
\multirow{5}{*}{MiniCPM4.1-8B} & GRPO@500 & +0.793 & +0.812 & +0.737 & \multirow{5}{*}{+0.732} \\
 & GRPO@200 & +0.601 & +0.588 & +0.540 &  \\
 & \cellcolor{grpC}SFT@100 & \cellcolor{grpC}+0.100 & \cellcolor{grpC}+0.426 & \cellcolor{grpC}+0.141 &  \\
 & \cellcolor{grpB}OPSD-LoRA@75 & \cellcolor{grpB}+0.336 & \cellcolor{grpB}+0.575 & \cellcolor{grpB}+0.294 &  \\
 & \cellcolor{grpC}OPSD@50 & \cellcolor{grpC}-0.269 & \cellcolor{grpC}+0.409 & \cellcolor{grpC}-0.018 &  \\
\bottomrule
\end{tabular}
\label{tab:question-rho}
\end{table}

\subsection{What the decomposition adds to the method comparison}
\label{app:methods-decomposition}

\Cref{app:methods} compares the three post-training methods by how far they drift
along $\hat v_{\mathrm{leak}}$. The decomposition lets us ask the same question in
units of accuracy.

On Qwen3-8B the checkpoints that gained and drifted onto the axis have
$L_{\mathrm{trained}}$ of $+0.123$ (GRPO@500), $+0.150$ (SFT@50) and $+0.121$
(OPSD@25). What the three methods buy along the leakage
direction is therefore comparable. Their $R_0$ is not: $+0.171$, $+0.089$ and
$+0.032$ over the same three checkpoints. The methods differ mainly in what they
achieved \emph{outside} the drift, and on-policy RL is the one that achieved most
of it.

This also explains why $\lambda$ orders the methods the way it does. OPSD@25 has
the largest $\lambda$ ($0.79$) not because it leaked more---its
$L_{\mathrm{trained}}$ is the smallest of the three---but because its total gain
is roughly half, so almost nothing it achieved lies outside the drift. A large
$\lambda$ can come from a large numerator or a small denominator, and here it is
the latter.

\subsection{An asymmetric response, and where post-training sits on it}
\label{app:response-asymmetry}
\Cref{sec:gain_audit} finds $\beta_{\mathrm{base}}$ and $\beta_{\mathrm{trained}}$
close to each other. Base counter-steering (\cref{app:counter-steer-base}) is the
one case that does not match: its slope is roughly half of either.

\begin{figure*}[h]
    \centering
    \includegraphics[width=0.75\textwidth]{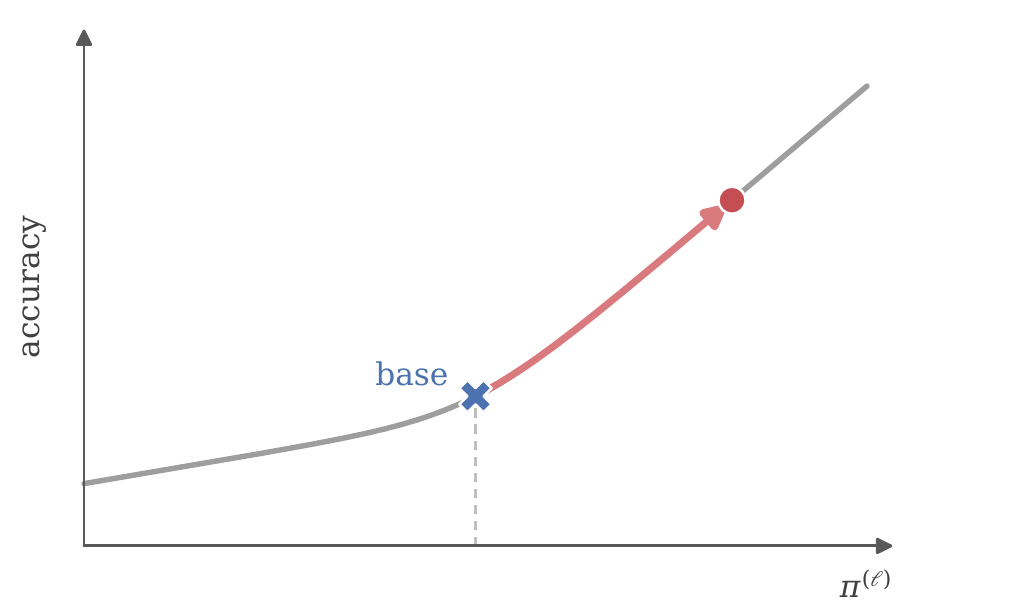}
    \caption{\textbf{A schematic showing that the response along $\hat{v}_{\text{leak}}$ is asymmetric about the base
model's position, and post-training moves the model along it.}}
\label{fig:response-schematic}
\end{figure*}

\Cref{fig:response-schematic} is the reading we favour. The accuracy response
along $\hat v_{\mathrm{leak}}$ is not a single slope but a curve, asymmetric about
the base model's own position: displacements above $\pi_0$ are worth about twice
as much as displacements below it, where the response compresses. Post-training
does not rebuild this curve, it moves the model along it. A checkpoint therefore
sits on the steep branch, which is why counter-steering a checkpoint and steering
the base model---both operating above $\pi_0$---return matching slopes, while
counter-steering the base does not.

Counter-steering at $\gamma=2$ supports the same picture
(\cref{app:counter-steer-x2}): it moves a checkpoint from $+\Delta\pi$ through
$\pi_0$ down to $-\Delta\pi$, so the second half of the trip lies in the
compressed region ($\pi < \pi_0$) and costs less than the first half. 

\subsection{Relation to three-way mediation decomposition}
\label{app:three-way}

The two paths in \cref{eq:bidirectional_gain_decomposition}
can be combined into a three-way accounting of the total gain:
\begin{equation}
\Delta
= \underbrace{R_0}_{\text{remaining gain}}
+ \underbrace{L_{\mathrm{base}}}_{\text{base steering effect}}
+ \underbrace{\mathcal{A}}_{\text{interaction}} .
\label{eq:three_way_gain_accounting}
\end{equation}
This parallels the algebraic structure of the three-way mediation
decomposition~\citep{vanderweele2013three}, with
$\mathcal{A}=L_{\mathrm{trained}}-L_{\mathrm{base}}$
capturing the difference in intervention effects between the base
and trained checkpoints.

When $\mathcal{A}$ is small, the two paths yield similar readings
of the gain.
Larger asymmetry would indicate that the effect of displacement
along $\hat v_{\mathrm{leak}}$ depends more strongly on the
checkpoint.
The three-way form makes this dependence explicit, providing
a useful extension for comparing models or training methods
with more divergent intervention responses.
We present this formulation as a theoretical extension of the
current decomposition~\cref{eq:bidirectional_gain_decomposition}, offering a framework for future studies
of how post-training changes the model's response to leakage.

\end{document}